\documentclass{article}

\usepackage{microtype}
\usepackage{graphicx}
\usepackage{subfigure}
\usepackage{booktabs} %
\usepackage{graphicx}
\usepackage{booktabs}
\usepackage{multirow}
\usepackage{diagbox}
\usepackage{arydshln}

\usepackage{hyperref}

\usepackage[accepted]{icml2024}

\usepackage{amsmath}
\usepackage{amssymb}
\usepackage{mathtools}
\usepackage{amsthm}

\usepackage[capitalize,noabbrev]{cleveref}

\theoremstyle{plain}

\theoremstyle{definition}

\theoremstyle{remark}

\usepackage{bigdelim}

\usepackage[textsize=tiny]{todonotes}

\newcommand{\tikzmark}[1]{\tikz[overlay,remember picture] \node (#1) {};}

\icmltitlerunning{The Entropy Enigma: Success and Failure of Entropy Minimization}

\begin{document}

\twocolumn[
\icmltitle{The Entropy Enigma: Success and Failure of Entropy Minimization}

\icmlsetsymbol{equal}{*}

\begin{icmlauthorlist}
\icmlauthor{Ori Press}{1}
\icmlauthor{Ravid Shwartz-Ziv}{2}
\icmlauthor{Yann LeCun}{2,3}
\icmlauthor{Matthias Bethge}{1}

\end{icmlauthorlist}

\icmlaffiliation{1}{University of T\"ubingen, T\"ubingen AI Center, Germany}
\icmlaffiliation{2}{New York University}
\icmlaffiliation{3}{Meta AI, FAIR}

\icmlcorrespondingauthor{}{ori.press@bethgelab.org}

\icmlkeywords{Machine Learning, ICML, entropy, test time adaptation, continual learning, accuracy estimation, out of distribution classification}

\vskip 0.3in
]

\printAffiliationsAndNotice{}  %

\begin{abstract}
Entropy minimization (EM) is frequently used to increase the accuracy of classification models when they're faced with new data at test time. EM is a self-supervised learning method that optimizes classifiers to assign even higher probabilities to their top predicted classes. In this paper, we analyze why EM works when adapting a model for a few steps and why it eventually fails after adapting for many steps. We show that, at first, EM causes the model to embed test images close to training images, thereby increasing model accuracy. After many steps of optimization, EM makes the model embed test images far away from the embeddings of training images, which results in a degradation of accuracy. Building upon our insights, we present a method for solving a practical problem: estimating a model's accuracy on a given arbitrary dataset without having access to its labels. Our method estimates accuracy by looking at how the embeddings of input images change as the model is optimized to minimize entropy. Experiments on 23 challenging datasets show that our method sets the SoTA with a mean absolute error of $5.75\%$, an improvement of $29.62\%$ over the previous SoTA on this task. Our code is available at: \url{https://github.com/oripress/EntropyEnigma} 
\end{abstract}

\section{Introduction}
Practitioners commonly employ model adaptation strategies to enhance classifier performance on real-world data, which often differs significantly from training data. Unsupervised losses play a crucial role in adapting models to images corrupted by noise, such as snow or motion blur, or images from domains not seen in training, such as paintings or computer rendered images. Entropy minimization (EM) is a Test Time Adaptation (TTA) method that can improve the accuracy of a model on new datasets, without the need for additional labeled training data. EM adapts classifiers by iteratively increasing the probabilities assigned to the most likely classes while diminishing those of the others, and is an integral part of many recent TTA methods \cite{wang2020tent, mummadi2021test, rusak2022if, goyal2022test, niu2022efficient, cho2023beyond, niu2023towards, press2023rdumb, dobler2024diversity, marsden2024universal}. In this paper, we analyze EM to understand how it works, when and why it fails, and how to use it to predict model accuracy.

The initial intuition behind using entropy minimization, given by \citet{wang2020tent} was based on the observation that models tend to be more accurate on images for which they make predictions with higher confidence. The logical extension of this observation was to encourage models to bolster their confidence on such images. However, our analysis reveals this intuition to be only partly true. Remarkably, even when we construct datasets by excluding samples initially classified correctly --- effectively creating datasets with a 100\% classification error rate at the start --- entropy minimization performance remains largely intact.

\begin{figure*}[!ht]
    \centering
    \begin{minipage}{0.7975\linewidth}
        \includegraphics[width=\linewidth]{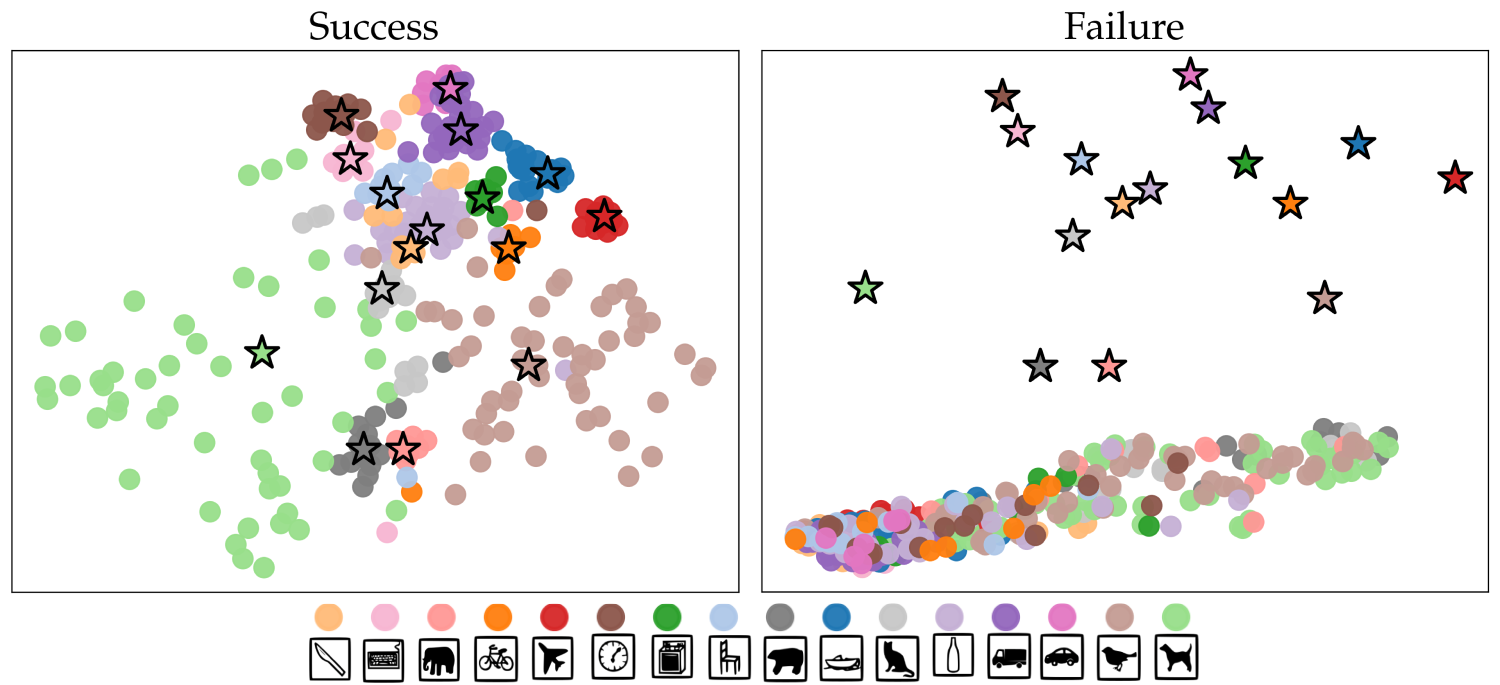}
    \end{minipage}
    \caption{\textbf{Understanding the successes and failures of EM through clustering embedding dynamics.} After a few iterations of adaptation (left), EM improves the accuracy of pretrained classifiers by embedding the input test data near mean embeddings of classes from the training data, marked by stars. Eventually, after many iterations (right), EM fails, because it embeds input test data far from where training data is embedded. We show the t-SNE embeddings of 16-class-Imagenet \cite{geirhos2018generalisation}, throughout adaptation to Gaussian Noise 3 \cite{hendrycks2019benchmarking}.}
    \label{fig:fig1}
\end{figure*}

Our analysis uncovers that during entropy minimization, embeddings of images from the input dataset tend to form distinct clusters. The distances between samples within each cluster diminish, creating more defined groupings, while the centers of these clusters gradually move apart, a phenomenon akin to neural collapse \cite{papyan2020prevalence, han2021neural, ben2023reverse}. At first, embeddings of the input images not only cluster, but also stay close to the embeddings of original training images. Only after numerous optimization steps do these embeddings begin to diverge, distancing themselves from the embeddings of the clean training data (Fig. \ref{fig:fig1}). We show this divergence to be intricately tied to a reduction in the model's accuracy.

Drawing from our insights, we present a method designed to estimate the accuracy of a given model on any dataset, without labels. This task is notably difficult, because in some cases in-distribution accuracy is tied to out-of-distribution (OOD) accuracy \cite{miller2021accuracy}, while in other cases it is not \cite{teney2022id}.  Our approach, termed Weighted Flips (WF), works in conjunction with TTA methods as they adapt to input data, with minimal added overhead. Using approximations of cluster consistency, WF estimates the accuracy of the network by measuring how the predictions of a fixed set of images change: the more they change, the lower the consistency of the clusters and the lower the predicted accuracy. We validate the efficacy of our method across an extensive array of 23 ImageNet-scale \cite{deng2009imagenet} datasets, encompassing diverse challenges, such as random adversarial noises, hard images, and datasets featuring OOD classes. WF surpasses the prior state-of-the-art methods by a substantial margin of 29.62\%, setting a new benchmark in the accuracy estimation domain.

\section{The Mystery of Entropy Minimization}

EM has been validated as effective in semi-supervised settings, with pioneering work by \cite{grandvalet2004semi} and subsequent advancements, such as Tent \cite{wang2020tent}, which demonstrated EM's ability to enhance the accuracy of pre-trained classifiers on unlabeled ImageNet-scale \cite{deng2009imagenet} datasets. EM operates by iteratively optimizing the model to minimize the entropy of the output classification probabilities, denoted by $H(\hat{y}) = -\sum_c p(\hat{y}_c) \log p(\hat{y}_c)$, where $\hat{y}$ is the logits vector and $p(\hat{y}_c)$ is the probability assigned to class $c$. This approach inherently boosts the likelihood of the most probable classes while diminishing that of the others.  \citet{wang2020tent} observed a correlation between lower output entropy and accuracy, indicating that images with low entropy outputs are more likely to be classified correctly. Subsequent studies, including \cite{niu2022efficient, press2023rdumb, marsden2024universal}, have built on this foundation, assigning more weight to lower-entropy samples, and even ignoring high-entropy samples entirely.

To assess the influence of correctly classified images on EM's effectiveness, we tested the effects of omitting images that were initially correctly classified  by the model. If such images are pivotal in EM's ability to enhance classifier performance, we expect a notable decline in the EM efficacy.

For this purpose, we utilized ImageNet-C \cite{hendrycks2019benchmarking} Gaussian Noise level 3, dividing it into training and holdout sets. The training set was replicated seven times, systematically omitting images for which the ground truth label lay somewhere in the pre-trained model's top-$k$ predictions, for ($k \in [1,2,3,5,10,20,50]$). Concretely, for $k=1$, all accurately classified images were excluded, and for $k=2$, images whose label ranked within the top two predictions were removed, and so forth. Each altered training set was used to adapt a Tented model. The model's accuracy was then evaluated on the holdout set, with evaluations every ten iterations, spanning a total of 1,000 iterations.

\begin{figure}[!h]
  \centering
  \includegraphics[width=0.95\linewidth]{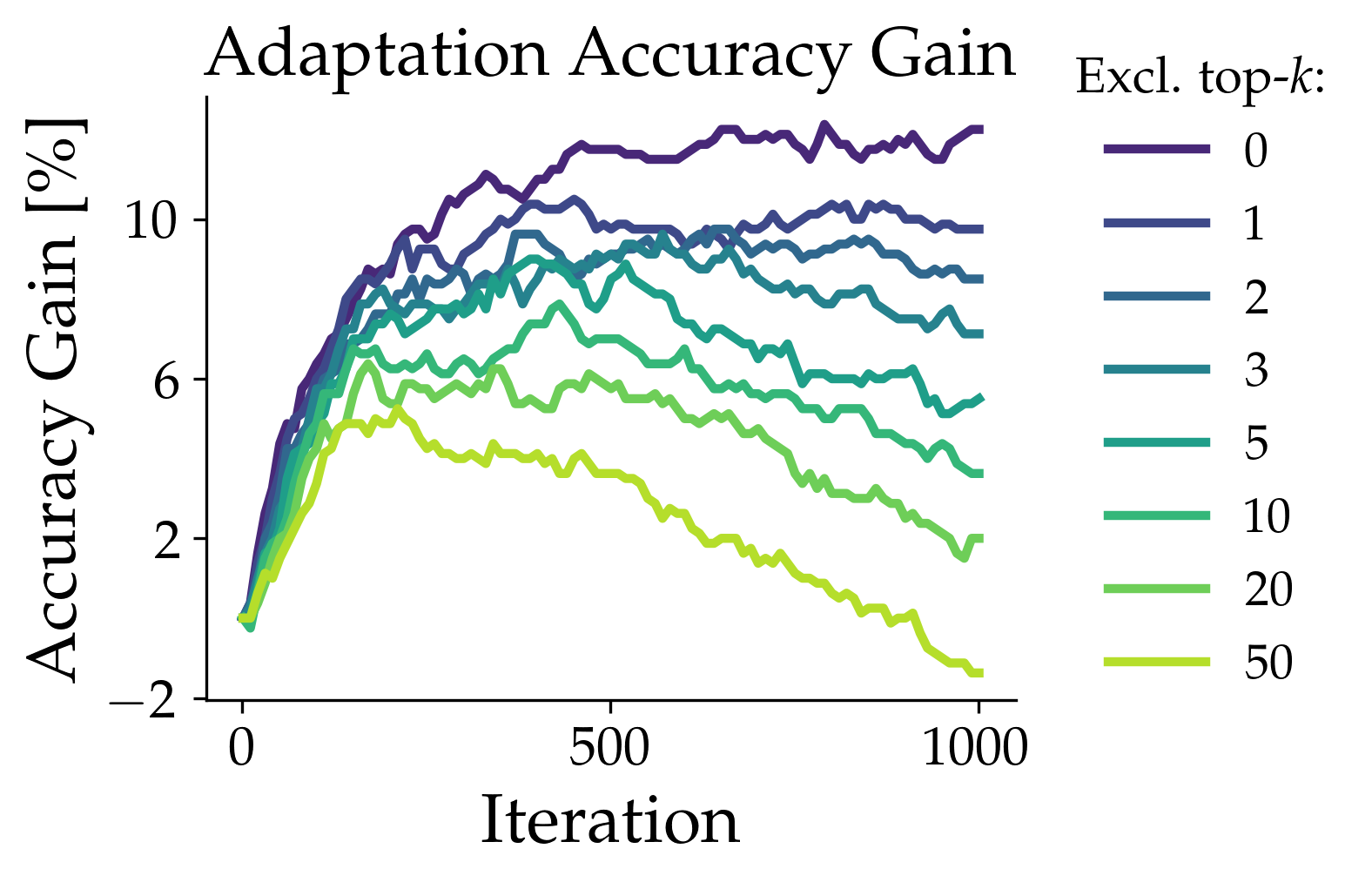}
  \label{fig:topk}

  \caption{\textbf{EM remains effective even when initially correctly classified images are excluded}. Accuracy gain per iteration on a holdout set, as Tent adapts to its inputs. Surprisingly, the performance gain on the holdout set is high, even when we exclude top-$k$ samples from the training set. When top-$k=0$, no images are excluded.}
\end{figure}

The experiment results (see Figure \ref{fig:topk}) are revealing, underscoring the robustness of EM. Notably, EM's effectiveness endures even when images initially classified correctly are excluded.

For instance, removing all initially correctly classified images before adaptation produces an increase in accuracy comparable to not removing any images, with gains of 10.50\% and 12.38\%, respectively. Even more remarkable is the persistence of this trend: with $k=10$, the model still registers a notable accuracy improvement of 7.88\%. This observation is particularly striking given the nature of the excluded images – they are not just numerous, but also represent the highest quality, being those the network is most certain about. Specifically, images excluded at $k=1$, which constitute 45\% of the dataset, have an average entropy of 1.85, markedly lower than the original dataset's average entropy of 2.84.

Additionally, we also tested the effects of removing images according on their initial entropy level, and found similar results (see Appendix \ref{appdx:omitting_samples}). These findings intriguingly suggest that the model's accuracy and entropy on individual images may not be as pivotal to EM's success in enhancing classifier performance as previously thought. It reveals a nuanced dimension of EM's functionality and hints at the presence of deeper mechanisms, which we will investigate next.

\section{Phases of Entropy Minimization: Clustering Dynamics and Embedding Alignment}

We analyze the evolution of input data embeddings as EM progresses through its iterations. At first, EM causes the model to increase in accuracy, which we refer to as the first phase, followed by a decrease in accuracy, which we refer to as the second phase. The number of EM iterations needed for the model to reach its maximum accuracy (the end of the first phase, and the beginning of the second) is varied and depends on the input data.

\begin{figure}[!h]
  \centering
  \includegraphics[width=0.95\linewidth]{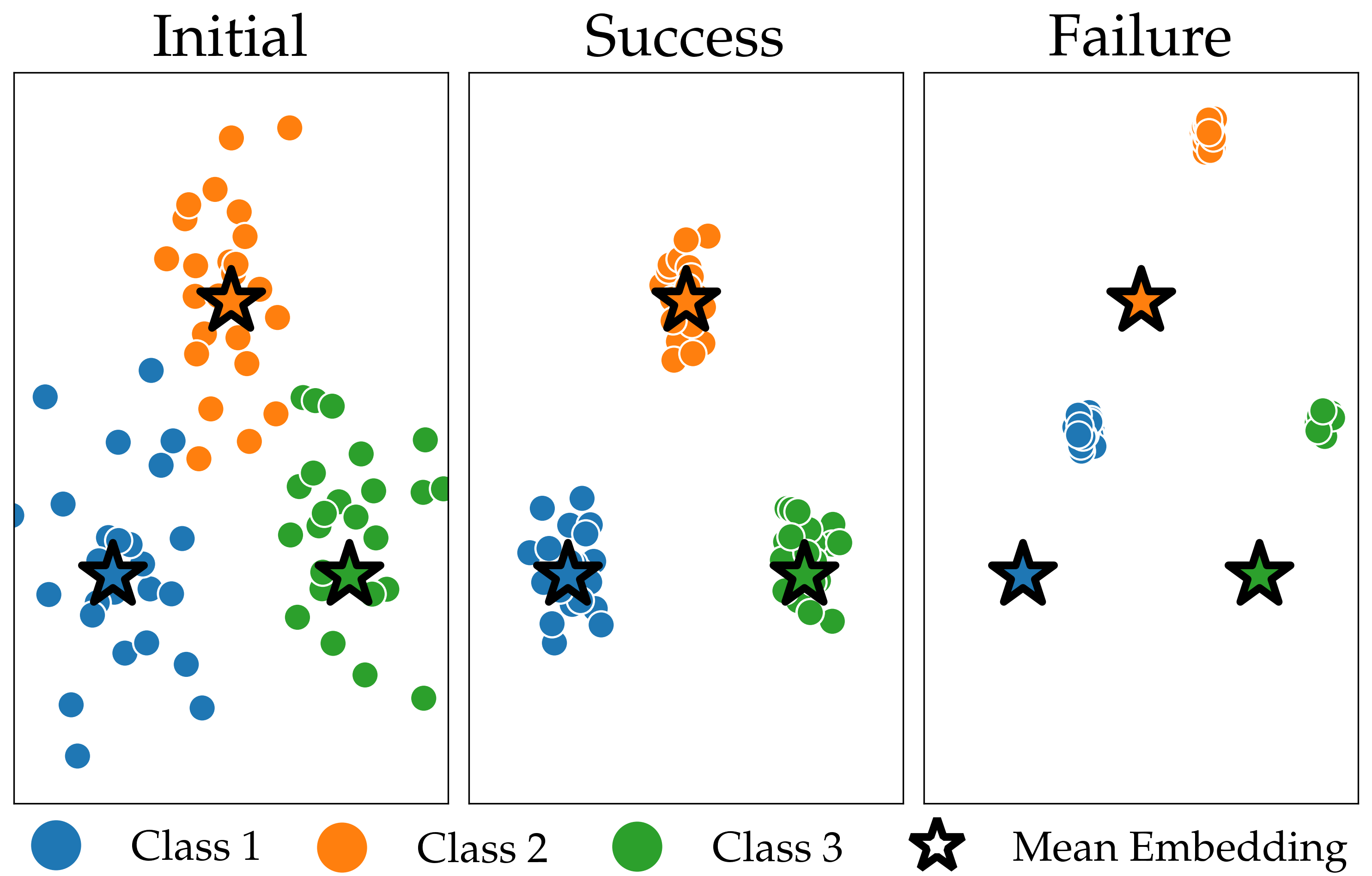}
  \caption{\textbf{The two-phase clustring paradigm explains EM behavior}. Intuitive visualization of EM's phases. In the first phase (success), input test data becomes more clustered, aligning closely with the mean embeddings of corresponding classes from the training data (the colored stars). In the second phase (failure), these clusters diverge from the mean embeddings.}
  \label{fig:clustering}
\end{figure}

In the first phase, these embeddings align closely with the embeddings of samples from the original training distribution. However, in the second phase, this alignment starts to deteriorate; the embeddings drift progressively further from the training distribution, disrupting the initial alignment, as conceptually depicted in Figure \ref{fig:clustering}.

To examine the clustering process across the two phases of the EM, we focus on two measures: (1) the quality of the clusters and (2) their alignment with the original training data distribution. For evaluating cluster quality, we ran k-means on the embeddings and computed the Silhouette score \cite{rousseeuw1987silhouettes}, a widely recognized metric for measuring cluster quality. The Silhouette score gauges how closely an embedding corresponds to its own cluster in contrast to neighboring clusters, with a high score indicating distinct and well-separated clusters.

To quantify the alignment between clusters and embeddings of the original training distribution, we
 looked at mean embeddings for the classes in the ImageNet validation set, alongside the centroids of clusters found by k-means. We use the Hungarian method \cite{kuhn1955hungarian} to find a matching between mean class embeddings and centroids, which minimizes the average distance between each assigned pair of (class embedding, centroid). Henceforth, we refer to this average of distances as ``Shift distance''.

As ImageNet contains many similar fine-grained classes, we restrict our analyses to the 16 classes outlined in \cite{geirhos2018generalisation}, which represent approximately 20\% of the total images. Consequently, we use $k=16$ when we cluster the embeddings using k-means. This focused approach allowed for a detailed and controlled examination of clustering behaviors within the framework of EM.

\begin{figure*}[!t]
    \centering
    \begin{minipage}{.49\linewidth}
        \centering
        \includegraphics[width=\linewidth]{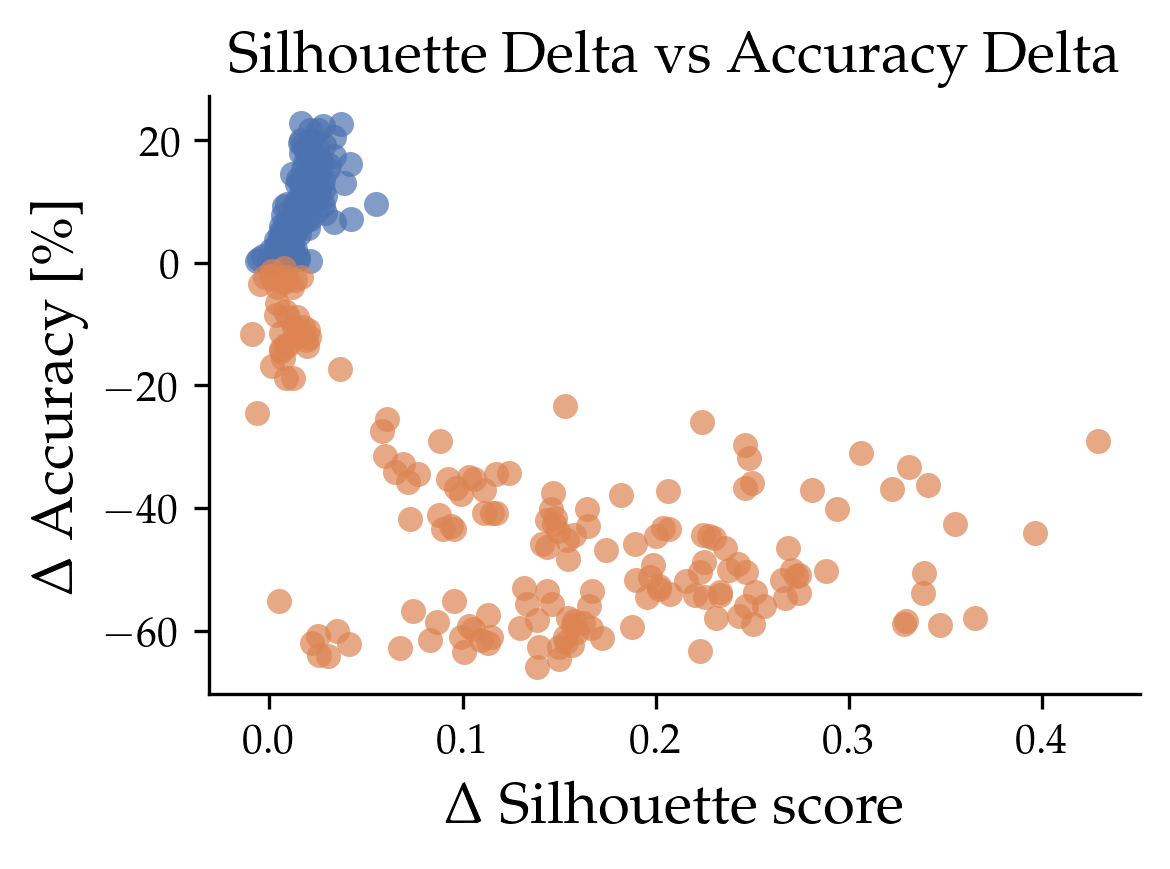}
    \end{minipage}
    \begin{minipage}{.49\linewidth}
        \centering

        \includegraphics[width=\linewidth]{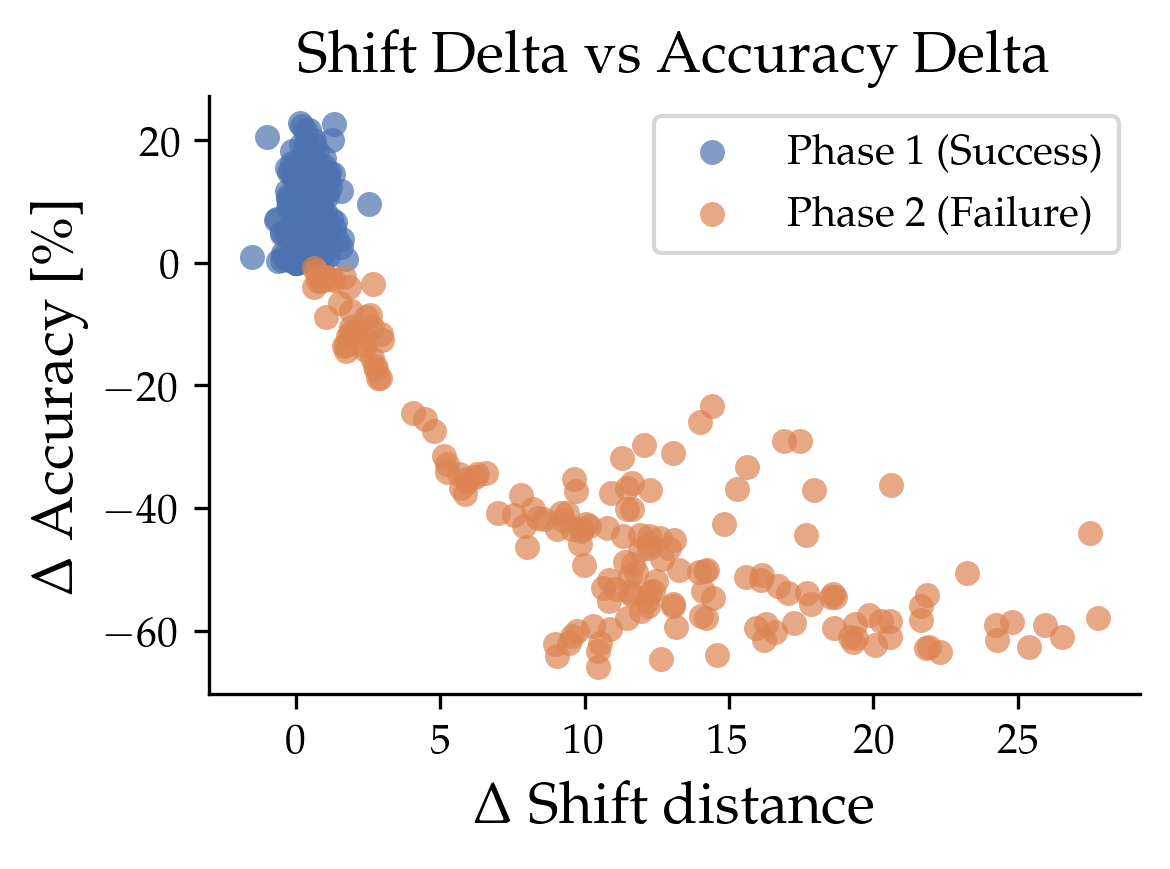}
    \end{minipage}
    \caption{\textbf{Two-phase behavior during the EM adaption predicts accuracy.} Differences in Silhouette score, Shift distance, and accuracy for Tent adaptation. Each point corresponds to a test dataset; each dataset appears twice: once in blue, corresponding to phase 1 (success, $\Delta$ Acc $\geq$ 0), and once in orange, corresponding to phase 2 (failure, $\Delta$ Acc $<$ 0). \textbf{Left:} In both phases, and across almost all datasets, the Silhouette score of embeddings increases, corresponding to a better-clustered embedding space.
    \textbf{Right:} In the first phase, input data embeddings are kept close to training image embeddings, while in the second phase, they drift away, exhibiting large Shift distance changes.  The datasets used are IN-C, IN-$\overline{\mbox{C}}$ and IN-3DCC.
    }
    \label{fig:silhouette}
\end{figure*}

We now examine changes in the Silhouette score and Shift distance as Tent adapts to the input data, over 50,000 iterations using a ResNet-50 \cite{he2016deep}. Figure \ref{fig:silhouette} showcases the comparative Silhouette scores and Shift distances for both phases, incorporating findings from three diverse datasets: IN-C \cite{hendrycks2019benchmarking}, IN-$\overline{\mbox{C}}$ \cite{mintun2021interaction}, and IN-3DCC \cite{kar20223d}.

Our findings distill into two primary insights: \textbf{First}, a positive change in Silhouette score, indicative of enhanced clustering, is observed in both phases for more than 98\% of cases. Notably, during the initial phase, a positive correlation exists between changes in Silhouette score and accuracy ($\rho = 0.70$, significant at $\alpha = 0.05$). \textbf{Second}, Shift distances minimally change (and sometimes diminish, signifying closer proximity to training data embeddings) in the first phase, they notably grow larger in the second phase. During this latter phase, a substantial negative correlation emerges between changes in Shift distance and accuracy ($\rho = -0.79$, significant at $\alpha = 0.05$).

Synthesizing these results reveals a nuanced picture: EM bolsters accuracy by clustering the embedded data into more concentrated clusters. This strategy remains efficacious as long as these embeddings align closely with the embeddings of the training data. However, as input data embeddings diverge from the training distribution, the classifier's accuracy diminishes. This intricate interplay offers a deeper understanding of EM's operation and its dependency on the spatial dynamics of data embeddings. We discuss the connection between EM and clustering in more detail in Appendix \ref{appdx:em_clustering}.

\section{Estimating Dataset Accuracy}
Leveraging our understanding of EM, we tackle a critical challenge in TTA settings:  estimating the accuracy of a classification model on a given dataset. Ideally, one might resort to the metrics used in this paper, namely Silhouette score or Shift distance, for this purpose. However, these metrics encounter practical hurdles: the Silhouette score depends on clustering, which varies across datasets due to differences in class distributions or the total number of classes, and calculating the Shift distance is impossible, as accessing the training data (in order to calculate mean embedding vectors per class) is forbidden in most TTA settings \cite{wang2020tent, niu2022efficient, yuan2023robust}.

\begin{figure}[!h]
  \centering
  \includegraphics[width=0.9995\linewidth]{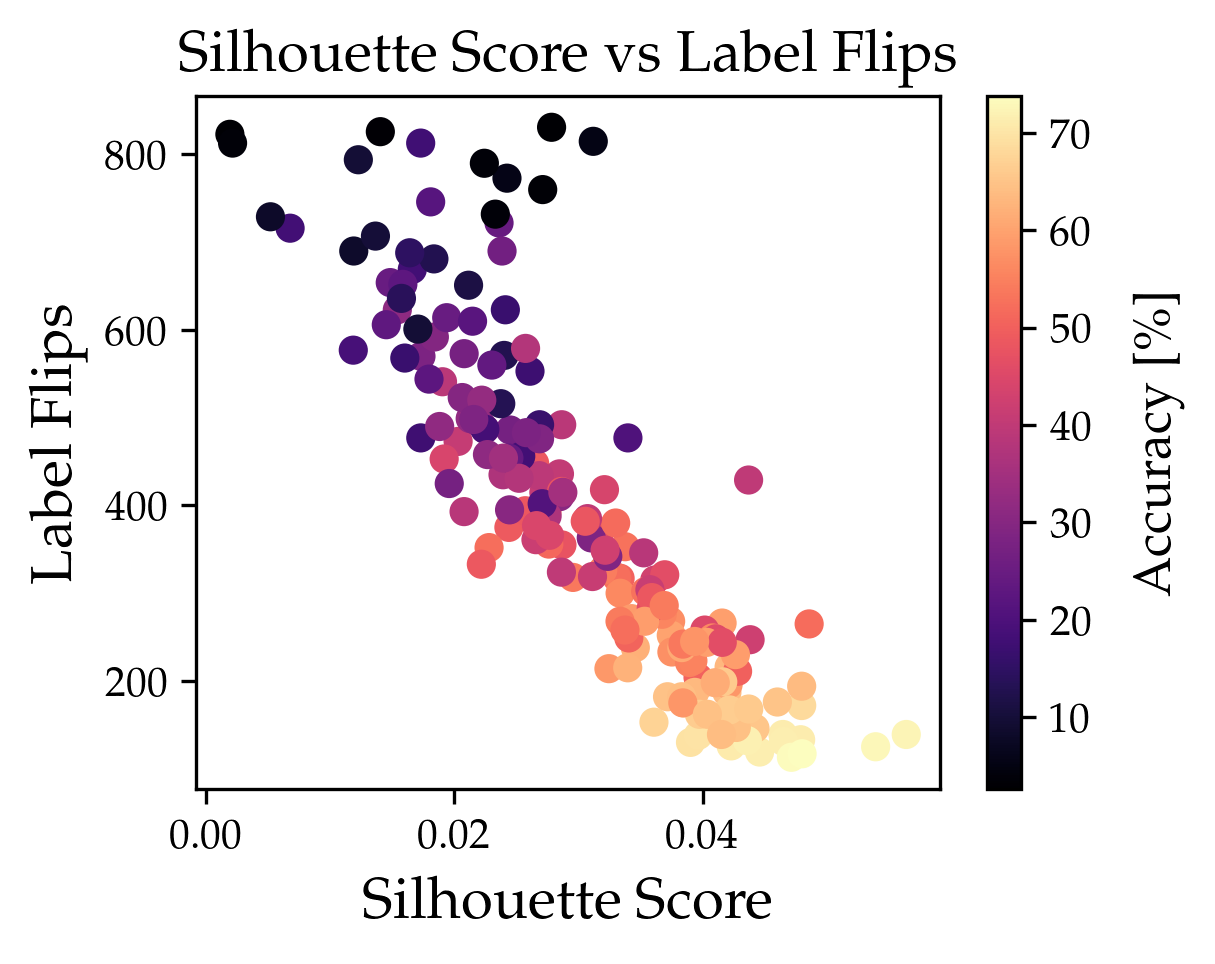}
  \caption{\textbf{Label flips are strongly correlated with Silhouette score}. Silhouette score at the initial iteration and the total number of label flips at the final iteration are correlated for datasets in IN-C, IN-$\overline{\mbox{C}}$, and IN-3DCC. Both metrics are correlated with accuracy, but measuring label flips is easier and more practical.}
  \label{fig:flips}
\end{figure}

\subsection{Label Flipping}

Due to the difficulties of measuring these scores in practice, we take a different approach. We look at the number of images for which the model's prediction changes somewhere between the initial and the final iteration of the EM (``label flips'').
According to our hypothesis, the number of label flips is correlated with the pre-trained model's accuracy on the dataset. Our reasoning is as follows: there exists a tight correlation between accuracy and Silhouette score at iteration 0 --- the higher the accuracy, the better clustered the input data, shown in Figure \ref{fig:flips}. Therefore, we do not expect EM, which works by clustering its inputs, to significantly change an already well-clustered set of embeddings. It follows that there will likely be only a few label flips. Conversely, given a dataset with a low  accuracy, its image embeddings will likely be badly clustered initially, which leads EM to change them significantly, resulting in many label flips.

We demonstrate the validity of this reasoning by adapting the state-of-the-art TTA method, Rdumb \cite{press2023rdumb}, to IN-C, IN-$\overline{\mbox{C}}$, and IN-3DCC. Initially, we used the pre-trained model to classify 1,000 input images and then recorded the total number of label flips after adaptation. The model is adapted for 1,000 iterations because Rdumb resets itself every 1,000 iterations. We find a strong correlation between accuracy and label flips, seen in Figure \ref{fig:flips}.

\subsection{Weighted Flips}
\begin{figure*}[!h]
    \centering
    \includegraphics[width=\linewidth]{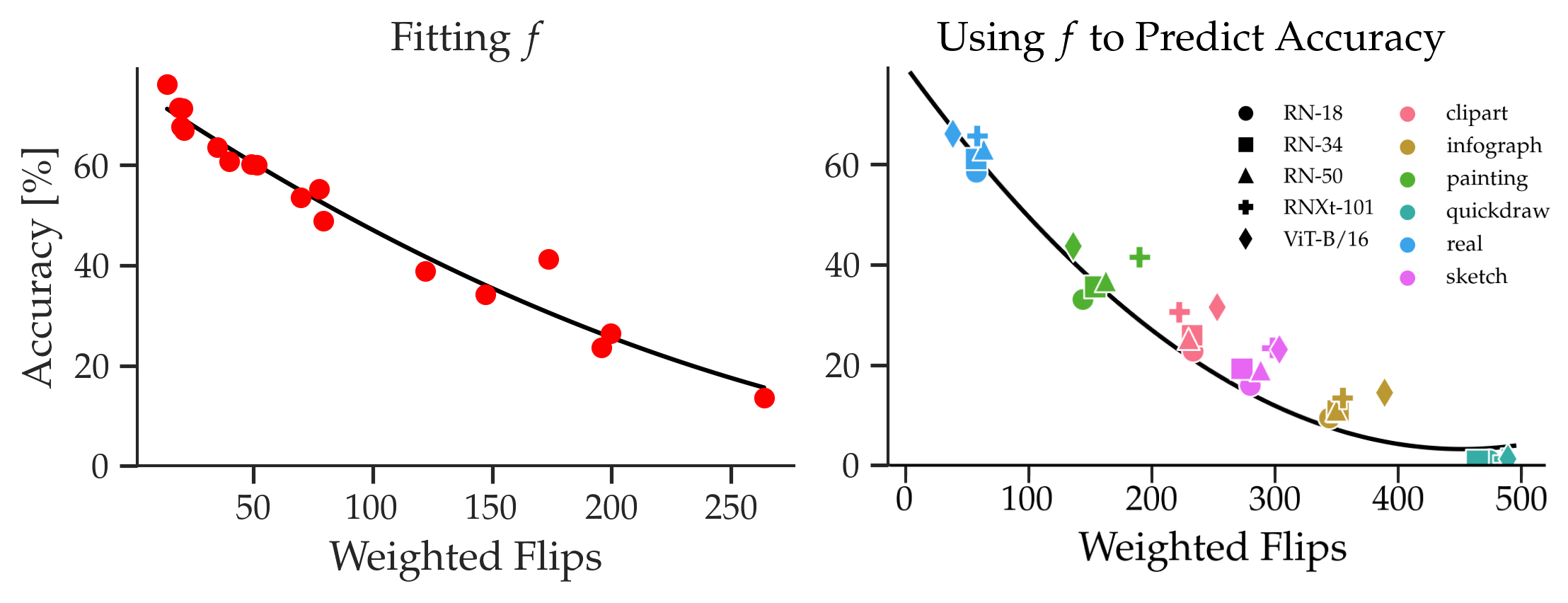}
    \caption{\textbf{Fitting and accuracy prediction using the WF method.} \textbf{Left:} Fitting $f$: using the noises in IN-C Holdout and ImageNet-Validation, we fit pairs of (weighted flips, accuracy), shown in red. The black curve shows the function resulting from interpolating the points, $f(x) = 0.00036 x^2 - 0.32 x + 75.66$.
    \textbf{Right:} With our weighted-flips-to-accuracy function $f$, we can estimate the accuracy of a model across the six splits from \cite{rusak2022imagenet}. We use the same $f$ function and show that it works across different architectures, without refitting.}
    \label{fig:estimating}
\end{figure*}

We now describe the Weighted Flips (WF) method of converting the count of label flips into a dataset accuracy estimate. Instead of just counting the number of flips, we additionally consider the classifier's initial confidence in its predictions for each image; images initially classified with high confidence that later flip should contribute more significantly than those with lower initial confidence. We then compute the WF as:

$$ WF = \sum_{i } 1_{\{flip\}}(i) \cdot c_i$$

where $1_{{flip}}(i)$ is 1 if image $i$'s label flipped and $0$ otherwise, and $c_i$ is the confidence percentile of image $i$. Utilizing pairs of weighted flips and accuracy (${(\text{WF}, \text{accuracy})}_k$) from IN-Validation and ImageNet-C holdout noises, we interpolate the weighted-flips-to-accuracy function, $f$ (refer to Figure \ref{fig:estimating}). To estimate the accuracy of a model on an unfamiliar dataset, we adapt the model to it using RDumb (for details, see Appendix \ref{appdx:rdumb}), measuring flips on the first 1,000 input images. After adaptation, we count and weigh the flips, estimating the model's accuracy as $f(\text{WF})$. Importantly, WF is versatile and can work with a range of TTA methods (see Appendix \ref{appdx:other_tta}), and $f$ can be interpolated in a variety of different ways (see Appendix \ref{appdx:f_abl}). In Appendix \ref{appdx:wf_iter}, \ref{appdx:wf_holdout}, we present ablation studies on the effects of varying end iterations and holdout set sizes on performance.

\subsection{Experimental Setting}
Accuracy estimation methods must yield robust estimates across diverse and challenging datasets to be considered reliable. In our evaluation, we probe the effectiveness of our proposed method using an extensive selection of popular ImageNet-scale classification datasets. This includes all classification datasets from the Shift-Happens benchmark\footnote{\footnotesize \url{https://github.com/shift-happens-benchmark/icml-2022}}. Our chosen datasets encompass a wide spectrum, from various types of noise (IN-C, IN-$\overline{\mbox{C}}$, IN-3DCC, CCC) and domain shifts (IN-R, IN-V2, IN-D), to adversarial noises (Patch-IN, BG Challenge, IN-Obfuscations), and even images featuring classes not present in ImageNet (NINCO).

Several datasets provide multiple splits of a similar nature, the results of which we average, except for ImageNet-D \cite{rusak2022imagenet}, which encompasses a variety of distinct domains. The CCC dataset \cite{press2023rdumb} is particularly expansive, containing 27 splits with 7.5M images each; for practicality, we only include the initial 25k images from each split in our analysis. Altogether, our evaluation spans 326 individual dataset splits.

We briefly describe the other methods tested alongside ours:

\begin{itemize}
\item AC \cite{hendrycks2016baseline}: Computes the dataset-wide average confidence for the top-predicted class in each image.
\item DoC \cite{guillory2021predicting}: Builds upon AC by assessing the variance in mean confidence between the validation and OOD sets, demonstrating consistent enhancements in performance.
\item ATC \cite{garg2022leveraging}: Estimates accuracy by determining the fraction of unlabeled data samples where the model's confidence exceeds a learned threshold.
\item COT \cite{lu2023predicting}: Estimates accuracy by applying Optimal Transport to quantify the disparity between OOD and in-distribution model outputs.
\end{itemize}

\subsection{Results}

\begin{table*}[!h]
    \centering
    \caption{Mean Absolute Error between estimated accuracy, and true accuracy on a ResNet-50 model, for 4 estimation methods (AC, DoC, ATC, COT) \cite{hendrycks2016baseline, guillory2021predicting,
    garg2022leveraging, lu2023predicting}, and ours. Our method (WF) is consistently either best or second best, with the best average and worst-case performance across many different OOD datasets. \textbf{Best} results are in bold; \underline{second best} are underlined, $\{.\}$ indicates how many splits are in each dataset, when there are more than 1.}

    \vskip 0.15in
    \begin{tabular}{cccccc}
        \toprule 
         Datasets
         & AC
         & DoC 
         & ATC 
         & COT  
         & WF (ours) \\
         \hline
         \multicolumn{1}{c}{\textit{Noises}} \\
         \hline
         IN-C  \{75\} \cite{hendrycks2019benchmarking} & 10.06 & 6.61 & 7.44 & \textbf{2.23}& \underline{4.79} \\
         IN-$\overline{\mbox{C}}$ \{50\} \cite{mintun2021interaction}  & 19.48 & 15.96 & 12.16 & \textbf{3.17} & \underline{7.35}  \\
         IN-3DCC \{60\} \cite{kar20223d} & 11.83 & \underline{3.44} & 8.15 & \textbf{3.02} & 3.66  \\
         CCC \{27\} \cite{press2023rdumb} & 15.51 & 11.95 & 6.05 & \textbf{2.04} & \underline{2.80} \\
         \hline
         \multicolumn{1}{c}{\textit{Domain Shifts}} \\
         \hline
        Stylized \cite{geirhos2018}  & 31.63 &  28.08 & 7.36 & \underline{12.18} & \textbf{3.81}  \\
         IN-V2 \{3\} \cite{recht2019imagenet}  & 5.58 & 2.41 & \textbf{0.45} & \underline{2.68}  & 4.70  \\
         IN-Sketch \cite{wang2019learning} & 22.34 & 18.78 & \textbf{0.15} & 4.23  & \underline{1.71}  \\
         IN-R \cite{hendrycks2021many} & 23.21 & 19.65 & \textbf{0.37} & 2.44  & \underline{1.88} \\
          IN-D \tikzmark{start}\cite{rusak2022imagenet} &  &  &  & &  \\
\hspace{15mm}    Real       &  10.56  &  7.00 &      \textbf{1.35}    &         27.54          & \underline{3.18}                           \\
\hspace{15mm} Painting   & 17.40 &  13.85 &    \underline{3.27}       &         7.49          &                     \textbf{2.12}                 \\
\hspace{15mm} Clipart    & 21.27   & 17.72    &   \textbf{1.62}       &         4.52         &                       \underline{3.37}               \\
\hspace{15mm} Sketch     & 24.43  &   20.87    &  \textbf{0.61}    &          \underline{0.71}         &                  5.44                    \\
\hspace{15mm} Infograph  & 54.12  &  50.57      &   36.26     &           \textbf{3.44}       &                         \underline{3.63}             \\
\hspace{15mm} Quickdraw  & 32.67  &  29.11      &   4.13    &            \textbf{1.60}       &   \underline{2.57}                                   \\
         Cartoon \& Drawing \{2\} \cite{salvador2022imagenet}  & 15.69 &  12.13 & \underline{4.42} & \textbf{1.62} & 13.25  \\

        \hline
        \multicolumn{1}{c}{\textit{Adversarial Noises}} \\
        \hline

        BG Challenge \{8\} \cite{xiao2020noise}  & 10.54 & 7.37 & \textbf{4.88} & 19.68  &  \underline{6.92}  \\
        IN-A \cite{hendrycks2021nae} & 45.12 & 41.57 & \textbf{20.51}  & 30.38 &  \underline{21.61}  \\
        IN-C Patch \{75\} \cite{gu2022evaluating}  & 4.37 & \textbf{0.16} & 4.42 & 2.57 & \underline{1.60}  \\
        IN-Hard \cite{taesiri2023zoom} & 29.71 & 26.15 & \underline{6.73} & 15.33 & \textbf{3.64} \\
        Patch-IN \{10\} \cite{pintor2023imagenet}  &  \underline{8.06} & \textbf{5.11} & \textbf{5.11} &  10.13 & 8.87  \\ 
        IN-Obfuscations \{3\} \cite{stimberg2023benchmarking}  & 99.90 & 96.34 & 99.90 & \textbf{0.12} & \underline{4.58}  \\
        \hline
        \multicolumn{1}{c}{\textit{OOD/Other}} \\
        \hline
        ObjectNet \cite{barbu2019objectnet} & 34.59 & 31.03 & \underline{9.43} & 10.40 & \textbf{2.74} \\
         NINCO \cite{bitterwolf2023or} & 50.29 & 46.74 & 26.97 & \underline{20.28} & \textbf{18.07}  \\
        \midrule
        \textbf{Average} & 26.02 & 22.29	& 11.81	& \underline{8.17}	& \textbf{5.75} \\
        \textbf{Worst Case} & 99.90 & 96.34 & 99.90 & \underline{30.38} & \textbf{21.61}  \\
        \textbf{Average (Worst Case Excluded)} & 22.66	& 18.92	& 7.81	& \underline{7.16}	& \textbf{5.03} \\
    \bottomrule
    \end{tabular}

\begin{tikzpicture}[overlay, remember picture, line width=1pt]

    \draw ([yshift=+0.1cm]start.south) -- ++(0,-2.55);

    \foreach \y in {-0.15,-0.6,-1.05,-1.5,-1.95,-2.4} {
        \draw[->] ([yshift=\y cm]start.south) -- ++(1,0);
    }
\end{tikzpicture}
\label{tbl:main}

\end{table*}

Looking at Table \ref{tbl:main} reveals that our WF method consistently stands out as the best estimator across a broad spectrum of ImageNet-scale datasets. WF sets a new benchmark by achieving an average estimation error of just 5.75\%, significantly outperforming the nearest competitor, COT, reducing the relative error by 29.62\%. This exemplary performance of WF is not limited to average cases; even in the most challenging scenarios of worst-case performance, WF maintains its superiority, cutting the error by 29.74\% compared to COT. Furthermore, WF demonstrates remarkable consistency as an estimator. In 18 of the 23 datasets evaluated, it either leads the pack or comes a close second. This is in stark contrast to the performance of COT, which, despite being second-best, only achieves top-two rankings in 12 datasets. The persistent effectiveness of WF across diverse conditions underscores its reliability and superiority in accuracy estimation.

\textbf{Practicality of WF:}
Beyond its top-tier performance, WF stands out for its practicality. It operates concurrently with the EM process, requiring only three parameters that define the weighted-flips-to-accuracy function, $f$. This process adds minimal computational overhead, requiring only 20 additional forward passes for every 1,000 Rdumb iteration steps. Lastly, WF is effective even when only a small number of samples are available, see Appendix \ref{appdx:wf_constraints}.

\begin{table*}[h] 
    \caption{
    Mean Absolute Error between estimated accuracy and true accuracy, across different architectures. Using the same weighted-flips to-accuracy function, $f$, works across different architectures and models, without need for finetuning. For each model and dataset, the task is to estimate the accuracy of that model on the dataset. \textbf{Best} results are in bold; \underline{second best} are underlined.  AugMix: $\diamondsuit$ ANT: $\ddagger$   DeepAugment: $\spadesuit$ \cite{hendrycks2019augmix, rusak2020simple, hendrycks2021many}}
    \vskip 0.15in

    \centering
        \begin{tabular}{c|c:ccccccccc}
        \toprule
    Datasets & \rotatebox{45}{RN-50} & \rotatebox{45}{RN-18} & \rotatebox{45}{RN-34} & \rotatebox{45}{RN-50 $\ddagger$} & \rotatebox{45}{RN-50 $\diamondsuit$} & \rotatebox{45}{RN-50 $\diamondsuit \spadesuit$} & \rotatebox{45}{RNXt-101} & \rotatebox{45}{RNXt-101 $\spadesuit$} & \rotatebox{45}{ViT-B/16} & \rotatebox{45}{MaxViT-T} \\
    \hline
        IN-C                     & \underline{4.79} & 7.21 & 6.04 & 5.39 & 5.02 & 4.81 & 5.35 & \textbf{4.12} & 8.34 & 6.73 \\ 
        IN-$\overline{\mbox{C}}$ & 7.35  & 7.90 & 6.77 & 6.84  & 6.60 & 6.48 & \textbf{5.60} & \underline{5.63} & 6.59 & 4.97 \\ 
        IN-3DCC                  & 3.66  & 3.58 & 3.89 & 3.20 & \underline{3.07} & \textbf{2.98} & 7.23 & 4.37 & 7.19  & 6.79\\ 
        IN-V2                    & 4.70  & 4.11 & \textbf{3.37} & \underline{3.67} & 5.06 & 5.00 & 6.47 & 5.54 & 4.44 & 6.08 \\
        
        \hspace{-5mm}   IN-D \begin{tikzpicture}
\draw[line width=0.5mm, ->] (-0.1,-0.2) -- (0.15,-0.2) -- (0.15,-0.45);
\end{tikzpicture} &  &  &  & & &  & & \\
        Real       & 3.18 & 2.83 & \textbf{0.38} & 2.72 & 6.59 & 3.30 & 4.24 & \underline{0.61} &  1.02   &  3.40              \\
        Painting    & 2.12 & 5.36 & 0.78 & \textbf{0.59} &  7.62  & 2.51 & 12.02 & 1.12 & 3.02  & \underline{0.60}         \\
        Clipart     & 3.37 & \underline{1.59} & 4.42 & 0.32 & 6.19 & 2.19 & 7.24 & \textbf{0.53} & 12.82  &  4.52              \\
        Sketch      & 5.44 & \textbf{1.53} & 3.93 & 6.18 & 9.73 & 3.60  & 10.75 & \underline{1.89}  &   11.04  & 10.88       \\
        Infograph  & 3.63 & \underline{1.76} & 3.67 & 3.74 & 6.78 & \textbf{0.28}  & 6.37 & 2.34 &    9.27 &  9.13          \\
        Quickdraw   & 2.57 & 2.34 & 2.24 & 2.20 & 2.53 & 1.27  & 2.27 & 1.21  &  2.36  &  2.31       \\
        \hline
        Average   & 4.08 & 3.82 & 3.55 & 3.49 & 5.92 & \underline{3.10} & 6.76 & \textbf{2.74} & 6.61 & 5.54  \\
        \bottomrule
        \end{tabular}
    
     \label{tab:models_and_arch}
\end{table*}

\textbf{Versatility across Models and Architectures:}
To demonstrate the adaptability of the WF method, we tested it across various models and architectures, employing the \textbf{same} weighted-flips-to-accuracy function, $f$, used in our primary experiments (Table \ref{tbl:main}). Testing encompassed different ResNet variants, including models enhanced with noise augmentation techniques, such as ANT \cite{rusak2020simple}, AugMix \cite{hendrycks2019augmix}, and DeepAugment \cite{hendrycks2021many}. Additionally, we evaluated a ResNext-101 \cite{xie2017aggregated}, ViTB-16 \cite{dosovitskiy2010image}, and MaxViT-T \cite{tu2022maxvit}. The mean absolute errors between estimated and actual accuracies are reported in Table \ref{tab:models_and_arch}. Remarkably, 5 of the 8 models tested achieved a lower mean absolute error than the baseline model, RN-50, showing that $f$ maintains its efficacy across different model architectures. When $f$ is refitted on the architecture that WF is evaluated on, performance improves (see Appendix \ref{appdx:vit}).

\begin{figure}[!h]
    \centering
    \includegraphics[width=0.98\linewidth]{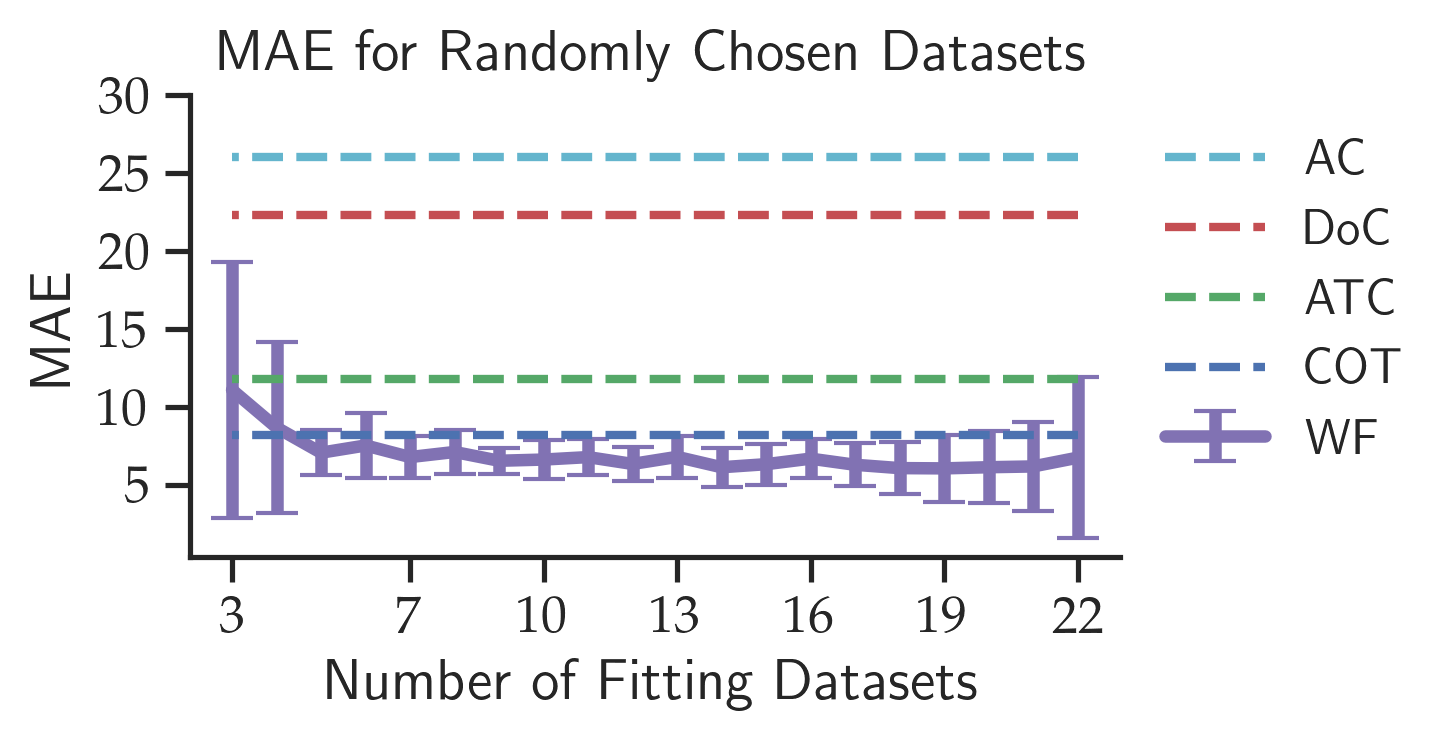}
    \caption{\textbf{WF outperforms other methods across almost all subset sizes}. Mean Absolute Error of WF when using a weighted-flips-to-accuracy function $f$ to fit on random subsets of the 23 datasets in Table \ref{tbl:main}. For each point on the x-axis, we sample 50 fitting datasets for WF, and plot the average and the standard deviation of the MAE. For the other methods, we plot average MAE across all datasets}.
    \label{fig:random-k}
\end{figure}

\textbf{Robustness to Dataset Choice:}
In Table \ref{tbl:main}, we derived the weighted-flips-to-accuracy function $f$ using IN-C holdout and ImageNet validation noises. We further validated the robustness of the WF method by fitting $f$ using a subset of the 23 datasets and then assessing its performance on the remaining datasets. As an added challenge, we excluded datasets used in the original configuration: IN-C Holdout and ImageNet-Validation. For each subset size, we repeated the fitting and evaluation process 50 times. The results, plotted in Figure \ref{fig:random-k}, illustrate that the WF method consistently outperforms COT across almost all subset sizes, reinforcing its resilience and reliability across a broad spectrum of datasets.

\section{Related Work}
To the best of our knowledge, the first time EM was shown to be useful for improving a classifier's accuracy was in  \cite{grandvalet2004semi}. They showed how EM can be applied to a logistic regressor, and found it to be beneficial in cases where the data was corrupted by outliers. Following this, \citet{lee2013pseudo} proposed pseudo labeling as a means of improving classification accuracy on MNIST. Interestingly, t-SNE is used to show that pseudo labeling works partly by encouraging the model's embeddings to be better clustered, and away from the decision boundaries of the model. Moreoever, it is stated that pseudo labeling is equivalent to entropy regularization \cite{grandvalet2004semi}. Although this might be true in the settings considered then, pseudo labeling was shown to be less effective (and thus not equivalent) on larger-scale datasets, by Tent. Unlike previous work, we demonstrate that EM clusters by measuring the Silhouette score of the clusters themselves, allowing us to empirically evaluate ImageNet scale datasets. Additionally, we show what happens when EM fails, which is not discussed in prior work, with the exception of \cite{oliver2018realistic}, which shows how EM fails to adapt to a toy ``two moons'' dataset, because the model increases the magnitude of its output logits. This isn't the case in most TTA settings, as the final layer of the model isn't trained.

Minimizing entropy at test time was popularized by Tent \cite{wang2020tent}, which demonstrated the effectiveness of EM on large-scale datasets, such as ImageNet-C.

Entropy minimization is ideal for domain adaptation: it can be used on a trained model, without retraining, and doesn't require balancing a proxy loss with a classification loss, as in \cite{gidaris2018unsupervised, sun2020test, gandelsman2022test}.

Though many prior works use losses that are based on entropy \cite{wang2020tent, rusak2022if, goyal2022test, mummadi2021test, wang2022continual, niu2022efficient,   cho2023beyond, press2023rdumb, niu2023towards, dobler2024diversity, marsden2024universal}, little is known as to \textit{why} it works. Additionally, entropy minimization, when used in TTA settings, is effective for only a limited number of iterations, before the classifier degrades to chance accuracy, shown in \cite{press2023rdumb}. Interestingly, this degradation of accuracy, named ``collapse'', differs from classical definitions of catastrophic forgetting in continual learning \cite{de2021continual}, in that the task itself does not change. 

A plethora of methods have been used for adapting a trained classifier to out-of-domain data: from using an auxiliary loss to help learn the test domain \cite{sun2019unsupervised, sun2020test, gandelsman2022test} through simply re-estimating the mean and variance statistics \cite{schneider2020improving, nado2020evaluating} to using image augmentations \cite{wang2022continual, song2023ecotta, chakrabarty2023santa}. However, for their simplicity and success, entropy minimization-based methods are still the most widely used and successful in settings most relevant to this work.

Works that follow Tent improve EM by modifying the loss to be more robust to label noise \cite{rusak2022if} or smoother \cite{mummadi2021test}, or by adjusting the temperature of the output distribution \cite{goyal2022test}. While testing on long sequences of images, both \cite{wang2022continual} and \cite{niu2022efficient} show that Tent degrades in accuracy, the more iterations it does. \cite{press2023rdumb} show that this is in fact true for all TTA methods apart from EATA \cite{niu2022efficient}, which uses an L2 regularizer to constrain the adapting model's weights to be close to those of the pretrained model. \cite{niu2023towards}
study the effects of batch size, label shifts and other factors on adaptation; they propose a method to stabilize adaptation. Similarly, \cite{dobler2024diversity} also test entropy minimization-based methods in real-world conditions, and propose a new method based on a diversity and a weighted entropy loss. Entropy has also been used in semi-supervised settings: \cite{sohn2020fixmatch} propose augmentation and an entropy loss to train a classifier when only a few labels are available.

Analyzing which labels flip during training has been studied in \cite{toneva2018empirical}, which explored which samples are forgotten during training. Another work, \cite{deng2022reducing} looked at how to reduce the amount of times a label flips during training. The agreement/disagreement between different models on ID data was shown to be linearly correlated to OOD accuracy and has been recently used to estimate accuracy in \cite{miller2021accuracy, jiang2021assessing, baek2022agreement, kim2023reliable}. These works are beyond the scope of this work, as they require access to multiple models and ID data, which is disallowed in most TTA settings \cite{wang2020tent, niu2022efficient, yuan2023robust}.

\section{Conclusion}

While EM is a cornerstone in many TTA methods, the mechanics of its success have remained enigmatic. This study sheds light on the transformative journey of input data embeddings under the  EM adaption. It reveals a biphasic clustering process, where alignment with the training data's embedding clusters bolsters accuracy, followed by a subsequent phase where excessive divergence diminishes it.

Our work goes beyond deciphering the mystery behind entropy minimization; it also utilizes this knowledge to significantly refine the precision of model accuracy predictions in TTA contexts. This dual achievement underscores the potential of deep analytical approaches in enhancing the efficacy and applicability of machine learning models.

\section*{Acknowledgements}
We thank Ofir Press for helpful insights and feedback. 

The authors thank the International Max Planck Research School for Intelligent Systems (IMPRS-IS) for supporting Ori Press. Matthias Bethge is a member of the Machine Learning Cluster of Excellence, funded by the Deutsche Forschungsgemeinschaft (DFG, German Research Foundation) under Germany’s Excellence Strategy – EXC number 2064/1 – Project number 390727645 and acknowledges support by the German Research Foundation (DFG): SFB 1233, Robust Vision: Inference Principles and Neural Mechanisms, TP 4, Project No: 276693517.
This work was supported by the Tübingen AI Center. The authors declare no conflicts of interests.

\section*{Impact Statement}
This paper presents work whose goal is to advance the field of Machine Learning. There are many potential societal consequences of our work, none which we feel needs to be highlighted here specifically.

\bibliography{bib.bib}

\begin{thebibliography}{72}
\providecommand{\natexlab}[1]{#1}
\providecommand{\url}[1]{\texttt{#1}}
\expandafter\ifx\csname urlstyle\endcsname\relax
  \providecommand{\doi}[1]{doi: #1}\else
  \providecommand{\doi}{doi: \begingroup \urlstyle{rm}\Url}\fi

\bibitem[Amini \& Gallinari(2002)Amini and Gallinari]{amini2002semi}
Amini, M.-R. and Gallinari, P.
\newblock Semi-supervised logistic regression.
\newblock In \emph{ECAI}, volume~2, pp.\ ~11, 2002.

\bibitem[Baek et~al.(2022)Baek, Jiang, Raghunathan, and
  Kolter]{baek2022agreement}
Baek, C., Jiang, Y., Raghunathan, A., and Kolter, J.~Z.
\newblock Agreement-on-the-line: Predicting the performance of neural networks
  under distribution shift.
\newblock \emph{Advances in Neural Information Processing Systems},
  35:\penalty0 19274--19289, 2022.

\bibitem[Barbu et~al.(2019)Barbu, Mayo, Alverio, Luo, Wang, Gutfreund,
  Tenenbaum, and Katz]{barbu2019objectnet}
Barbu, A., Mayo, D., Alverio, J., Luo, W., Wang, C., Gutfreund, D., Tenenbaum,
  J., and Katz, B.
\newblock Objectnet: A large-scale bias-controlled dataset for pushing the
  limits of object recognition models.
\newblock \emph{Advances in neural information processing systems}, 32, 2019.

\bibitem[Ben-Shaul et~al.(2023)Ben-Shaul, Shwartz-Ziv, Galanti, Dekel, and
  LeCun]{ben2023reverse}
Ben-Shaul, I., Shwartz-Ziv, R., Galanti, T., Dekel, S., and LeCun, Y.
\newblock Reverse engineering self-supervised learning.
\newblock \emph{arXiv preprint arXiv:2305.15614}, 2023.

\bibitem[Bitterwolf et~al.(2023)Bitterwolf, M{\"u}ller, and
  Hein]{bitterwolf2023or}
Bitterwolf, J., M{\"u}ller, M., and Hein, M.
\newblock In or out? fixing imagenet out-of-distribution detection evaluation.
\newblock \emph{arXiv preprint arXiv:2306.00826}, 2023.

\bibitem[Chakrabarty et~al.(2023)Chakrabarty, Sreenivas, and
  Biswas]{chakrabarty2023santa}
Chakrabarty, G., Sreenivas, M., and Biswas, S.
\newblock Santa: Source anchoring network and target alignment for continual
  test time adaptation.
\newblock \emph{Transactions on Machine Learning Research}, 2023.

\bibitem[Cho et~al.(2023)Cho, Kim, and Lee]{cho2023beyond}
Cho, Y., Kim, Y., and Lee, D.
\newblock Beyond entropy: Style transfer guided single image continual
  test-time adaptation.
\newblock \emph{arXiv preprint arXiv:2311.18270}, 2023.

\bibitem[De~Lange et~al.(2021)De~Lange, Aljundi, Masana, Parisot, Jia,
  Leonardis, Slabaugh, and Tuytelaars]{de2021continual}
De~Lange, M., Aljundi, R., Masana, M., Parisot, S., Jia, X., Leonardis, A.,
  Slabaugh, G., and Tuytelaars, T.
\newblock A continual learning survey: Defying forgetting in classification
  tasks.
\newblock \emph{IEEE transactions on pattern analysis and machine
  intelligence}, 44\penalty0 (7):\penalty0 3366--3385, 2021.

\bibitem[Dempster et~al.(1977)Dempster, Laird, and Rubin]{dempster1977maximum}
Dempster, A.~P., Laird, N.~M., and Rubin, D.~B.
\newblock Maximum likelihood from incomplete data via the em algorithm.
\newblock \emph{Journal of the royal statistical society: series B
  (methodological)}, 39\penalty0 (1):\penalty0 1--22, 1977.

\bibitem[Deng et~al.(2009)Deng, Dong, Socher, Li, Li, and
  Fei-Fei]{deng2009imagenet}
Deng, J., Dong, W., Socher, R., Li, L.-J., Li, K., and Fei-Fei, L.
\newblock Imagenet: A large-scale hierarchical image database.
\newblock In \emph{2009 IEEE conference on computer vision and pattern
  recognition}, pp.\  248--255. Ieee, 2009.

\bibitem[Deng et~al.(2022)Deng, Xiao, Long, and Zhang]{deng2022reducing}
Deng, X., Xiao, Y., Long, B., and Zhang, Z.
\newblock Reducing flipping errors in deep neural networks.
\newblock In \emph{Proceedings of the AAAI Conference on Artificial
  Intelligence}, volume~36, pp.\  6506--6514, 2022.

\bibitem[D{\"o}bler et~al.(2024)D{\"o}bler, Marencke, Marsden, and
  Yang]{dobler2024diversity}
D{\"o}bler, M., Marencke, F., Marsden, R.~A., and Yang, B.
\newblock Diversity-aware buffer for coping with temporally correlated data
  streams in online test-time adaptation.
\newblock \emph{arXiv preprint arXiv:2401.00989}, 2024.

\bibitem[Dosovitskiy et~al.(2010)Dosovitskiy, Beyer, Kolesnikov, Weissenborn,
  Zhai, Unterthiner, Dehghani, Minderer, Heigold, Gelly,
  et~al.]{dosovitskiy2010image}
Dosovitskiy, A., Beyer, L., Kolesnikov, A., Weissenborn, D., Zhai, X.,
  Unterthiner, T., Dehghani, M., Minderer, M., Heigold, G., Gelly, S., et~al.
\newblock An image is worth 16x16 words: Transformers for image recognition at
  scale. arxiv 2020.
\newblock \emph{arXiv preprint arXiv:2010.11929}, 2010.

\bibitem[Gandelsman et~al.(2022)Gandelsman, Sun, Chen, and
  Efros]{gandelsman2022test}
Gandelsman, Y., Sun, Y., Chen, X., and Efros, A.
\newblock Test-time training with masked autoencoders.
\newblock \emph{Advances in Neural Information Processing Systems},
  35:\penalty0 29374--29385, 2022.

\bibitem[Garg et~al.(2022)Garg, Balakrishnan, Lipton, Neyshabur, and
  Sedghi]{garg2022leveraging}
Garg, S., Balakrishnan, S., Lipton, Z.~C., Neyshabur, B., and Sedghi, H.
\newblock Leveraging unlabeled data to predict out-of-distribution performance.
\newblock \emph{arXiv preprint arXiv:2201.04234}, 2022.

\bibitem[Geirhos et~al.(2018)Geirhos, Temme, Rauber, Sch{\"u}tt, Bethge, and
  Wichmann]{geirhos2018generalisation}
Geirhos, R., Temme, C.~R., Rauber, J., Sch{\"u}tt, H.~H., Bethge, M., and
  Wichmann, F.~A.
\newblock Generalisation in humans and deep neural networks.
\newblock \emph{Advances in neural information processing systems}, 31, 2018.

\bibitem[Geirhos et~al.(2019)Geirhos, Rubisch, Michaelis, Bethge, Wichmann, and
  Brendel]{geirhos2018}
Geirhos, R., Rubisch, P., Michaelis, C., Bethge, M., Wichmann, F.~A., and
  Brendel, W.
\newblock Imagenet-trained {CNN}s are biased towards texture; increasing shape
  bias improves accuracy and robustness.
\newblock In \emph{International Conference on Learning Representations}, 2019.
\newblock URL \url{https://openreview.net/forum?id=Bygh9j09KX}.

\bibitem[Gidaris et~al.(2018)Gidaris, Singh, and
  Komodakis]{gidaris2018unsupervised}
Gidaris, S., Singh, P., and Komodakis, N.
\newblock Unsupervised representation learning by predicting image rotations.
\newblock \emph{arXiv preprint arXiv:1803.07728}, 2018.

\bibitem[Goyal et~al.(2022)Goyal, Sun, Raghunathan, and Kolter]{goyal2022test}
Goyal, S., Sun, M., Raghunathan, A., and Kolter, J.~Z.
\newblock Test time adaptation via conjugate pseudo-labels.
\newblock \emph{Advances in Neural Information Processing Systems},
  35:\penalty0 6204--6218, 2022.

\bibitem[Grandvalet \& Bengio(2004)Grandvalet and Bengio]{grandvalet2004semi}
Grandvalet, Y. and Bengio, Y.
\newblock Semi-supervised learning by entropy minimization.
\newblock \emph{Advances in neural information processing systems}, 17, 2004.

\bibitem[Gu et~al.(2022)Gu, Tresp, and Qin]{gu2022evaluating}
Gu, J., Tresp, V., and Qin, Y.
\newblock Evaluating model robustness to patch perturbations.
\newblock In \emph{ICML 2022 Shift Happens Workshop}, 2022.

\bibitem[Guillory et~al.(2021)Guillory, Shankar, Ebrahimi, Darrell, and
  Schmidt]{guillory2021predicting}
Guillory, D., Shankar, V., Ebrahimi, S., Darrell, T., and Schmidt, L.
\newblock Predicting with confidence on unseen distributions.
\newblock In \emph{Proceedings of the IEEE/CVF International Conference on
  Computer Vision}, pp.\  1134--1144, 2021.

\bibitem[Han et~al.(2021)Han, Papyan, and Donoho]{han2021neural}
Han, X., Papyan, V., and Donoho, D.~L.
\newblock Neural collapse under mse loss: Proximity to and dynamics on the
  central path.
\newblock \emph{arXiv preprint arXiv:2106.02073}, 2021.

\bibitem[He et~al.(2016)He, Zhang, Ren, and Sun]{he2016deep}
He, K., Zhang, X., Ren, S., and Sun, J.
\newblock Deep residual learning for image recognition.
\newblock In \emph{Proceedings of the IEEE conference on computer vision and
  pattern recognition}, pp.\  770--778, 2016.

\bibitem[Hendrycks \& Dietterich(2019)Hendrycks and
  Dietterich]{hendrycks2019benchmarking}
Hendrycks, D. and Dietterich, T.
\newblock Benchmarking neural network robustness to common corruptions and
  perturbations.
\newblock \emph{arXiv preprint arXiv:1903.12261}, 2019.

\bibitem[Hendrycks \& Gimpel(2016)Hendrycks and Gimpel]{hendrycks2016baseline}
Hendrycks, D. and Gimpel, K.
\newblock A baseline for detecting misclassified and out-of-distribution
  examples in neural networks.
\newblock \emph{arXiv preprint arXiv:1610.02136}, 2016.

\bibitem[Hendrycks et~al.(2019)Hendrycks, Mu, Cubuk, Zoph, Gilmer, and
  Lakshminarayanan]{hendrycks2019augmix}
Hendrycks, D., Mu, N., Cubuk, E.~D., Zoph, B., Gilmer, J., and
  Lakshminarayanan, B.
\newblock Augmix: A simple data processing method to improve robustness and
  uncertainty.
\newblock \emph{arXiv preprint arXiv:1912.02781}, 2019.

\bibitem[Hendrycks et~al.(2021{\natexlab{a}})Hendrycks, Basart, Mu, Kadavath,
  Wang, Dorundo, Desai, Zhu, Parajuli, Guo, et~al.]{hendrycks2021many}
Hendrycks, D., Basart, S., Mu, N., Kadavath, S., Wang, F., Dorundo, E., Desai,
  R., Zhu, T., Parajuli, S., Guo, M., et~al.
\newblock The many faces of robustness: A critical analysis of
  out-of-distribution generalization.
\newblock In \emph{Proceedings of the IEEE/CVF International Conference on
  Computer Vision}, pp.\  8340--8349, 2021{\natexlab{a}}.

\bibitem[Hendrycks et~al.(2021{\natexlab{b}})Hendrycks, Zhao, Basart,
  Steinhardt, and Song]{hendrycks2021nae}
Hendrycks, D., Zhao, K., Basart, S., Steinhardt, J., and Song, D.
\newblock Natural adversarial examples.
\newblock \emph{CVPR}, 2021{\natexlab{b}}.

\bibitem[Ioffe \& Szegedy(2015)Ioffe and Szegedy]{ioffe2015batch}
Ioffe, S. and Szegedy, C.
\newblock Batch normalization: Accelerating deep network training by reducing
  internal covariate shift.
\newblock In \emph{International conference on machine learning}, pp.\
  448--456. pmlr, 2015.

\bibitem[Jiang et~al.(2021)Jiang, Nagarajan, Baek, and
  Kolter]{jiang2021assessing}
Jiang, Y., Nagarajan, V., Baek, C., and Kolter, J.~Z.
\newblock Assessing generalization of sgd via disagreement.
\newblock \emph{arXiv preprint arXiv:2106.13799}, 2021.

\bibitem[Kar et~al.(2022)Kar, Yeo, Atanov, and Zamir]{kar20223d}
Kar, O.~F., Yeo, T., Atanov, A., and Zamir, A.
\newblock 3d common corruptions and data augmentation.
\newblock In \emph{Proceedings of the IEEE/CVF Conference on Computer Vision
  and Pattern Recognition}, pp.\  18963--18974, 2022.

\bibitem[Kim et~al.(2023)Kim, Sun, Raghunathan, and Kolter]{kim2023reliable}
Kim, E., Sun, M., Raghunathan, A., and Kolter, Z.
\newblock Reliable test-time adaptation via agreement-on-the-line.
\newblock \emph{arXiv preprint arXiv:2310.04941}, 2023.

\bibitem[Krizhevsky et~al.(2009)Krizhevsky, Hinton,
  et~al.]{krizhevsky2009learning}
Krizhevsky, A., Hinton, G., et~al.
\newblock Learning multiple layers of features from tiny images.
\newblock 2009.

\bibitem[Kuhn(1955)]{kuhn1955hungarian}
Kuhn, H.~W.
\newblock The hungarian method for the assignment problem.
\newblock \emph{Naval research logistics quarterly}, 2\penalty0 (1-2):\penalty0
  83--97, 1955.

\bibitem[Kullback \& Leibler(1951)Kullback and
  Leibler]{kullback1951information}
Kullback, S. and Leibler, R.~A.
\newblock On information and sufficiency.
\newblock \emph{The annals of mathematical statistics}, 22\penalty0
  (1):\penalty0 79--86, 1951.

\bibitem[Lee et~al.(2013)]{lee2013pseudo}
Lee, D.-H. et~al.
\newblock Pseudo-label: The simple and efficient semi-supervised learning
  method for deep neural networks.
\newblock In \emph{Workshop on challenges in representation learning, ICML},
  volume~3, pp.\  896. Atlanta, 2013.

\bibitem[Lu et~al.(2023)Lu, Wang, Zhai, Kolouri, Campbell, and
  Sycara]{lu2023predicting}
Lu, Y., Wang, Z., Zhai, R., Kolouri, S., Campbell, J., and Sycara, K.
\newblock Predicting out-of-distribution error with confidence optimal
  transport.
\newblock \emph{arXiv preprint arXiv:2302.05018}, 2023.

\bibitem[Marsden et~al.(2024)Marsden, D{\"o}bler, and
  Yang]{marsden2024universal}
Marsden, R.~A., D{\"o}bler, M., and Yang, B.
\newblock Universal test-time adaptation through weight ensembling, diversity
  weighting, and prior correction.
\newblock In \emph{Proceedings of the IEEE/CVF Winter Conference on
  Applications of Computer Vision}, pp.\  2555--2565, 2024.

\bibitem[Miller et~al.(2021)Miller, Taori, Raghunathan, Sagawa, Koh, Shankar,
  Liang, Carmon, and Schmidt]{miller2021accuracy}
Miller, J.~P., Taori, R., Raghunathan, A., Sagawa, S., Koh, P.~W., Shankar, V.,
  Liang, P., Carmon, Y., and Schmidt, L.
\newblock Accuracy on the line: on the strong correlation between
  out-of-distribution and in-distribution generalization.
\newblock In \emph{International Conference on Machine Learning}, pp.\
  7721--7735. PMLR, 2021.

\bibitem[Mintun et~al.(2021)Mintun, Kirillov, and Xie]{mintun2021interaction}
Mintun, E., Kirillov, A., and Xie, S.
\newblock On interaction between augmentations and corruptions in natural
  corruption robustness.
\newblock \emph{Advances in Neural Information Processing Systems},
  34:\penalty0 3571--3583, 2021.

\bibitem[Mummadi et~al.(2021)Mummadi, Hutmacher, Rambach, Levinkov, Brox, and
  Metzen]{mummadi2021test}
Mummadi, C.~K., Hutmacher, R., Rambach, K., Levinkov, E., Brox, T., and Metzen,
  J.~H.
\newblock Test-time adaptation to distribution shift by confidence maximization
  and input transformation.
\newblock \emph{arXiv preprint arXiv:2106.14999}, 2021.

\bibitem[Nado et~al.(2020)Nado, Padhy, Sculley, D'Amour, Lakshminarayanan, and
  Snoek]{nado2020evaluating}
Nado, Z., Padhy, S., Sculley, D., D'Amour, A., Lakshminarayanan, B., and Snoek,
  J.
\newblock Evaluating prediction-time batch normalization for robustness under
  covariate shift.
\newblock \emph{arXiv preprint arXiv:2006.10963}, 2020.

\bibitem[Niu et~al.(2022)Niu, Wu, Zhang, Chen, Zheng, Zhao, and
  Tan]{niu2022efficient}
Niu, S., Wu, J., Zhang, Y., Chen, Y., Zheng, S., Zhao, P., and Tan, M.
\newblock Efficient test-time model adaptation without forgetting.
\newblock In \emph{International conference on machine learning}, pp.\
  16888--16905. PMLR, 2022.

\bibitem[Niu et~al.(2023)Niu, Wu, Zhang, Wen, Chen, Zhao, and
  Tan]{niu2023towards}
Niu, S., Wu, J., Zhang, Y., Wen, Z., Chen, Y., Zhao, P., and Tan, M.
\newblock Towards stable test-time adaptation in dynamic wild world.
\newblock \emph{arXiv preprint arXiv:2302.12400}, 2023.

\bibitem[Oliver et~al.(2018)Oliver, Odena, Raffel, Cubuk, and
  Goodfellow]{oliver2018realistic}
Oliver, A., Odena, A., Raffel, C., Cubuk, E., and Goodfellow, I.
\newblock Realistic evaluation of semi-supervised learning algortihms.
\newblock In \emph{International conference on learning representations}, pp.\
  1--15, 2018.

\bibitem[Papyan et~al.(2020)Papyan, Han, and Donoho]{papyan2020prevalence}
Papyan, V., Han, X., and Donoho, D.~L.
\newblock Prevalence of neural collapse during the terminal phase of deep
  learning training.
\newblock \emph{Proceedings of the National Academy of Sciences}, 117\penalty0
  (40):\penalty0 24652--24663, 2020.

\bibitem[Pintor et~al.(2023)Pintor, Angioni, Sotgiu, Demetrio, Demontis,
  Biggio, and Roli]{pintor2023imagenet}
Pintor, M., Angioni, D., Sotgiu, A., Demetrio, L., Demontis, A., Biggio, B.,
  and Roli, F.
\newblock Imagenet-patch: A dataset for benchmarking machine learning
  robustness against adversarial patches.
\newblock \emph{Pattern Recognition}, 134:\penalty0 109064, 2023.

\bibitem[Poland \& Shachter(1993)Poland and Shachter]{poland1993mixtures}
Poland, W.~B. and Shachter, R.~D.
\newblock Mixtures of gaussians and minimum relative entropy techniques for
  modeling continuous uncertainties.
\newblock In \emph{Uncertainty in Artificial Intelligence}, pp.\  183--190.
  Elsevier, 1993.

\bibitem[Press et~al.(2023)Press, Schneider, K{\"u}mmerer, and
  Bethge]{press2023rdumb}
Press, O., Schneider, S., K{\"u}mmerer, M., and Bethge, M.
\newblock Rdumb: A simple approach that questions our progress in continual
  test-time adaptation.
\newblock \emph{Advances in Neural Information Processing Systems}, 36, 2023.

\bibitem[Recht et~al.(2019)Recht, Roelofs, Schmidt, and
  Shankar]{recht2019imagenet}
Recht, B., Roelofs, R., Schmidt, L., and Shankar, V.
\newblock Do imagenet classifiers generalize to imagenet?
\newblock In \emph{International conference on machine learning}, pp.\
  5389--5400. PMLR, 2019.

\bibitem[Rousseeuw(1987)]{rousseeuw1987silhouettes}
Rousseeuw, P.~J.
\newblock Silhouettes: a graphical aid to the interpretation and validation of
  cluster analysis.
\newblock \emph{Journal of computational and applied mathematics}, 20:\penalty0
  53--65, 1987.

\bibitem[Rusak et~al.(2020)Rusak, Schott, Zimmermann, Bitterwolf, Bringmann,
  Bethge, and Brendel]{rusak2020simple}
Rusak, E., Schott, L., Zimmermann, R.~S., Bitterwolf, J., Bringmann, O.,
  Bethge, M., and Brendel, W.
\newblock A simple way to make neural networks robust against diverse image
  corruptions.
\newblock In \emph{Computer Vision--ECCV 2020: 16th European Conference,
  Glasgow, UK, August 23--28, 2020, Proceedings, Part III 16}, pp.\  53--69.
  Springer, 2020.

\bibitem[Rusak et~al.(2022{\natexlab{a}})Rusak, Schneider, Gehler, Bringmann,
  Brendel, and Bethge]{rusak2022imagenet}
Rusak, E., Schneider, S., Gehler, P.~V., Bringmann, O., Brendel, W., and
  Bethge, M.
\newblock Imagenet-d: A new challenging robustness dataset inspired by domain
  adaptation.
\newblock In \emph{ICML 2022 Shift Happens Workshop}, 2022{\natexlab{a}}.

\bibitem[Rusak et~al.(2022{\natexlab{b}})Rusak, Schneider, Pachitariu, Eck,
  Gehler, Bringmann, Brendel, and Bethge]{rusak2022if}
Rusak, E., Schneider, S., Pachitariu, G., Eck, L., Gehler, P.~V., Bringmann,
  O., Brendel, W., and Bethge, M.
\newblock If your data distribution shifts, use self-learning.
\newblock \emph{Transactions on Machine Learning Research}, 2022{\natexlab{b}}.

\bibitem[Salvador \& Oberman(2022)Salvador and Oberman]{salvador2022imagenet}
Salvador, T. and Oberman, A.~M.
\newblock Imagenet-cartoon and imagenet-drawing: two domain shift datasets for
  imagenet.
\newblock In \emph{ICML 2022 Shift Happens Workshop}, 2022.

\bibitem[Schneider et~al.(2020)Schneider, Rusak, Eck, Bringmann, Brendel, and
  Bethge]{schneider2020improving}
Schneider, S., Rusak, E., Eck, L., Bringmann, O., Brendel, W., and Bethge, M.
\newblock Improving robustness against common corruptions by covariate shift
  adaptation.
\newblock \emph{Advances in neural information processing systems},
  33:\penalty0 11539--11551, 2020.

\bibitem[Sohn et~al.(2020)Sohn, Berthelot, Carlini, Zhang, Zhang, Raffel,
  Cubuk, Kurakin, and Li]{sohn2020fixmatch}
Sohn, K., Berthelot, D., Carlini, N., Zhang, Z., Zhang, H., Raffel, C.~A.,
  Cubuk, E.~D., Kurakin, A., and Li, C.-L.
\newblock Fixmatch: Simplifying semi-supervised learning with consistency and
  confidence.
\newblock \emph{Advances in neural information processing systems},
  33:\penalty0 596--608, 2020.

\bibitem[Song et~al.(2023)Song, Lee, Kweon, and Choi]{song2023ecotta}
Song, J., Lee, J., Kweon, I.~S., and Choi, S.
\newblock Ecotta: Memory-efficient continual test-time adaptation via
  self-distilled regularization.
\newblock In \emph{Proceedings of the IEEE/CVF Conference on Computer Vision
  and Pattern Recognition}, pp.\  11920--11929, 2023.

\bibitem[Stimberg et~al.(2023)Stimberg, Chakrabarti, Lu, Hazimeh, Stretcu,
  Qiao, Liu, Kaya, Rashtchian, Fuxman, et~al.]{stimberg2023benchmarking}
Stimberg, F., Chakrabarti, A., Lu, C.-T., Hazimeh, H., Stretcu, O., Qiao, W.,
  Liu, Y., Kaya, M., Rashtchian, C., Fuxman, A., et~al.
\newblock Benchmarking robustness to adversarial image obfuscations.
\newblock \emph{arXiv preprint arXiv:2301.12993}, 2023.

\bibitem[Sun et~al.(2019)Sun, Tzeng, Darrell, and Efros]{sun2019unsupervised}
Sun, Y., Tzeng, E., Darrell, T., and Efros, A.~A.
\newblock Unsupervised domain adaptation through self-supervision.
\newblock \emph{arXiv preprint arXiv:1909.11825}, 2019.

\bibitem[Sun et~al.(2020)Sun, Wang, Liu, Miller, Efros, and Hardt]{sun2020test}
Sun, Y., Wang, X., Liu, Z., Miller, J., Efros, A., and Hardt, M.
\newblock Test-time training with self-supervision for generalization under
  distribution shifts.
\newblock In \emph{International conference on machine learning}, pp.\
  9229--9248. PMLR, 2020.

\bibitem[Taesiri et~al.(2023)Taesiri, Nguyen, Habchi, Bezemer, and
  Nguyen]{taesiri2023zoom}
Taesiri, M.~R., Nguyen, G., Habchi, S., Bezemer, C.-P., and Nguyen, A.
\newblock Zoom is what you need: An empirical study of the power of zoom and
  spatial biases in image classification.
\newblock \emph{arXiv preprint arXiv:2304.05538}, 2023.

\bibitem[Teney et~al.(2022)Teney, Lin, Oh, and Abbasnejad]{teney2022id}
Teney, D., Lin, Y., Oh, S.~J., and Abbasnejad, E.
\newblock Id and ood performance are sometimes inversely correlated on
  real-world datasets.
\newblock \emph{arXiv preprint arXiv:2209.00613}, 2022.

\bibitem[Toneva et~al.(2018)Toneva, Sordoni, Combes, Trischler, Bengio, and
  Gordon]{toneva2018empirical}
Toneva, M., Sordoni, A., Combes, R. T.~d., Trischler, A., Bengio, Y., and
  Gordon, G.~J.
\newblock An empirical study of example forgetting during deep neural network
  learning.
\newblock \emph{arXiv preprint arXiv:1812.05159}, 2018.

\bibitem[Tu et~al.(2022)Tu, Talebi, Zhang, Yang, Milanfar, Bovik, and
  Li]{tu2022maxvit}
Tu, Z., Talebi, H., Zhang, H., Yang, F., Milanfar, P., Bovik, A., and Li, Y.
\newblock Maxvit: Multi-axis vision transformer.
\newblock In \emph{European conference on computer vision}, pp.\  459--479.
  Springer, 2022.

\bibitem[Wang et~al.(2020)Wang, Shelhamer, Liu, Olshausen, and
  Darrell]{wang2020tent}
Wang, D., Shelhamer, E., Liu, S., Olshausen, B., and Darrell, T.
\newblock Tent: Fully test-time adaptation by entropy minimization.
\newblock \emph{arXiv preprint arXiv:2006.10726}, 2020.

\bibitem[Wang et~al.(2019)Wang, Ge, Lipton, and Xing]{wang2019learning}
Wang, H., Ge, S., Lipton, Z., and Xing, E.~P.
\newblock Learning robust global representations by penalizing local predictive
  power.
\newblock In \emph{Advances in Neural Information Processing Systems}, pp.\
  10506--10518, 2019.

\bibitem[Wang et~al.(2022)Wang, Fink, Van~Gool, and Dai]{wang2022continual}
Wang, Q., Fink, O., Van~Gool, L., and Dai, D.
\newblock Continual test-time domain adaptation.
\newblock In \emph{Proceedings of the IEEE/CVF Conference on Computer Vision
  and Pattern Recognition}, pp.\  7201--7211, 2022.

\bibitem[Xiao et~al.(2020)Xiao, Engstrom, Ilyas, and Madry]{xiao2020noise}
Xiao, K., Engstrom, L., Ilyas, A., and Madry, A.
\newblock Noise or signal: The role of image backgrounds in object recognition.
\newblock \emph{ArXiv preprint arXiv:2006.09994}, 2020.

\bibitem[Xie et~al.(2017)Xie, Girshick, Doll{\'a}r, Tu, and
  He]{xie2017aggregated}
Xie, S., Girshick, R., Doll{\'a}r, P., Tu, Z., and He, K.
\newblock Aggregated residual transformations for deep neural networks.
\newblock In \emph{Proceedings of the IEEE conference on computer vision and
  pattern recognition}, pp.\  1492--1500, 2017.

\bibitem[Yuan et~al.(2023)Yuan, Xie, and Li]{yuan2023robust}
Yuan, L., Xie, B., and Li, S.
\newblock Robust test-time adaptation in dynamic scenarios.
\newblock In \emph{Proceedings of the IEEE/CVF Conference on Computer Vision
  and Pattern Recognition}, pp.\  15922--15932, 2023.

\end{thebibliography}
\bibliographystyle{icml2024}

\newpage
\appendix
\onecolumn

\section{The Relationship between Entropy Minimization and Clustering}
\label{appdx:em_clustering}

In this section, we explain the connection between entropy minimization and the Expectation-Maximization algorithm \cite{dempster1977maximum} with a mixture of Gaussians and show how the iterative entropy minimization objective leads to a clustering process similar to the Expectation-Maximization algorithm.

In the Expectation-Maximization algorithm for clustering, the latent variables represent the cluster assignments, and the algorithm alternates between estimating the cluster assignments (E-step) and updating the cluster parameters (M-step). The convergence of the EM algorithm in this setting has been formally established \cite{dempster1977maximum}.

\citet{poland1993mixtures} showed that for a random variable $X$ with a given distribution and the mixture of random variables $Y$ that derive from it, the objective of minimizing the ``relative entropy'' between $X$ and $Y$ generalizes the objective of the Expectation Maximization algorithm: to maximize the likelihood of the observations $x$ drawn from $Y$'s distribution.

In our setting, the iterative entropy minimization process corresponds to the Expectation Maximization algorithm, as iterative entropy minimization can also be seen as a form of ``self-training'' with minimization of the relative entropy (the DKL \cite{kullback1951information}) of the pseudo-labels (the model's predictions) \cite{grandvalet2004semi}. The forward pass of our training process serves two purposes: (1) it sets the ``observations'',  which are the model's predictions, and (2) it acts as the E-step of the algorithm, estimating the distribution given the model parameters (the clustering assignment). The backpropagation step, which updates the model parameters (the cluster parameters), serves as the M-step and maximizes the likelihood under the current pseudo-label estimates \cite{amini2002semi}. It is important to note that in our setting, the entropy minimization procedure involves changing both $X$ and $Y$ in each iteration, which may be different from the original Expectation Maximization algorithm.

Using these insights, we can provide a better explanation for the two-phase clustering phenomenon observed in our experiments. In the initial ``success'' phase, where the change in the embeddings is relatively small during the process, the entropy minimization effectively performs Expectation Maximization unsupervised clustering in the model's embedding space, guided by the smart initialization provided by the pre-trained model. The E-step estimates the pseudo-labels based on the current embedding structure, while the M-step updates the model to refine the embeddings and increase intra-cluster similarity. This process leads to the formation of well-separated clusters, as reflected by the increasing Silhouette score.

However, as the Expectation Maximization algorithm continues over many iterations in the ``failure'' phase or if there is bad initialization, it starts to overfit the model to the specific characteristics of the "new" test data. Unlike the regular Expectation Maximization algorithm, in our case, the data distribution (the observations) changes over time, which leads to a drift in the embeddings away from the initialized representations learned from the training data. This overfitting effect, which might even converge to a global minimum, is captured by the increasing Shift distance between the test data embeddings and the training class embeddings.

\begin{figure}[h!]
    \centering
    \includegraphics[width=0.8\linewidth]{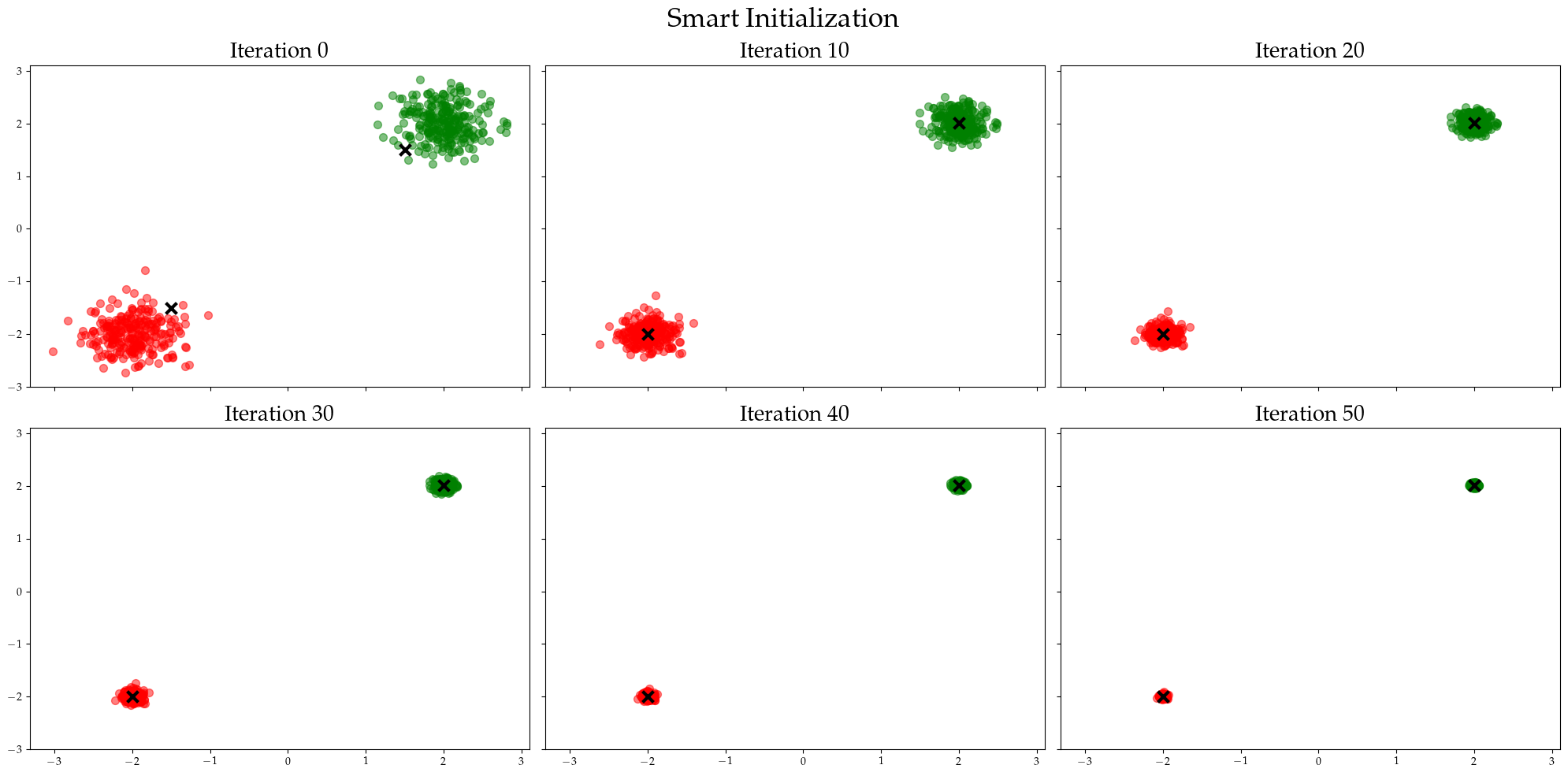}
    
    \includegraphics[width=0.8\linewidth]{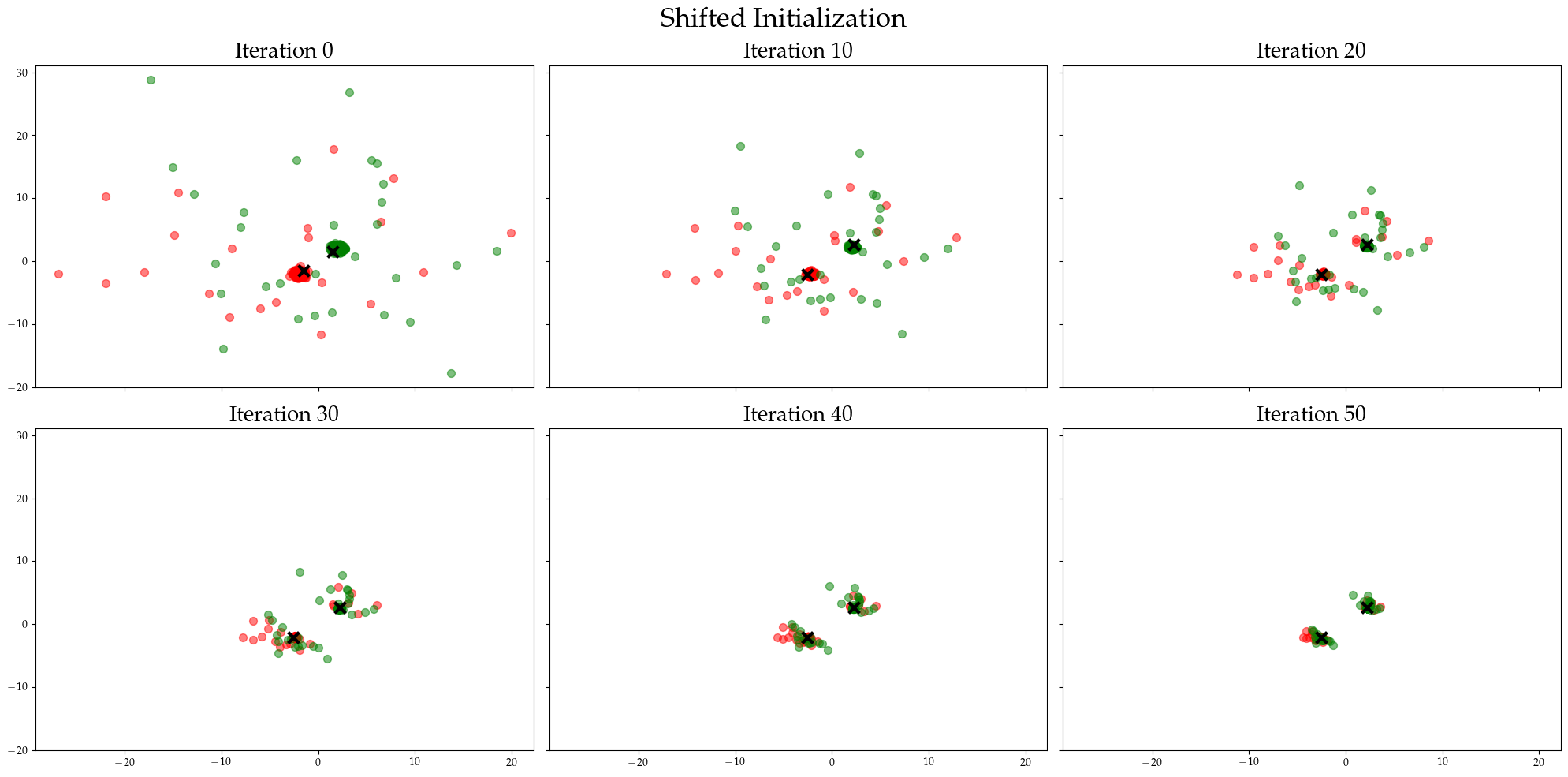}

    \includegraphics[width=0.8\linewidth]{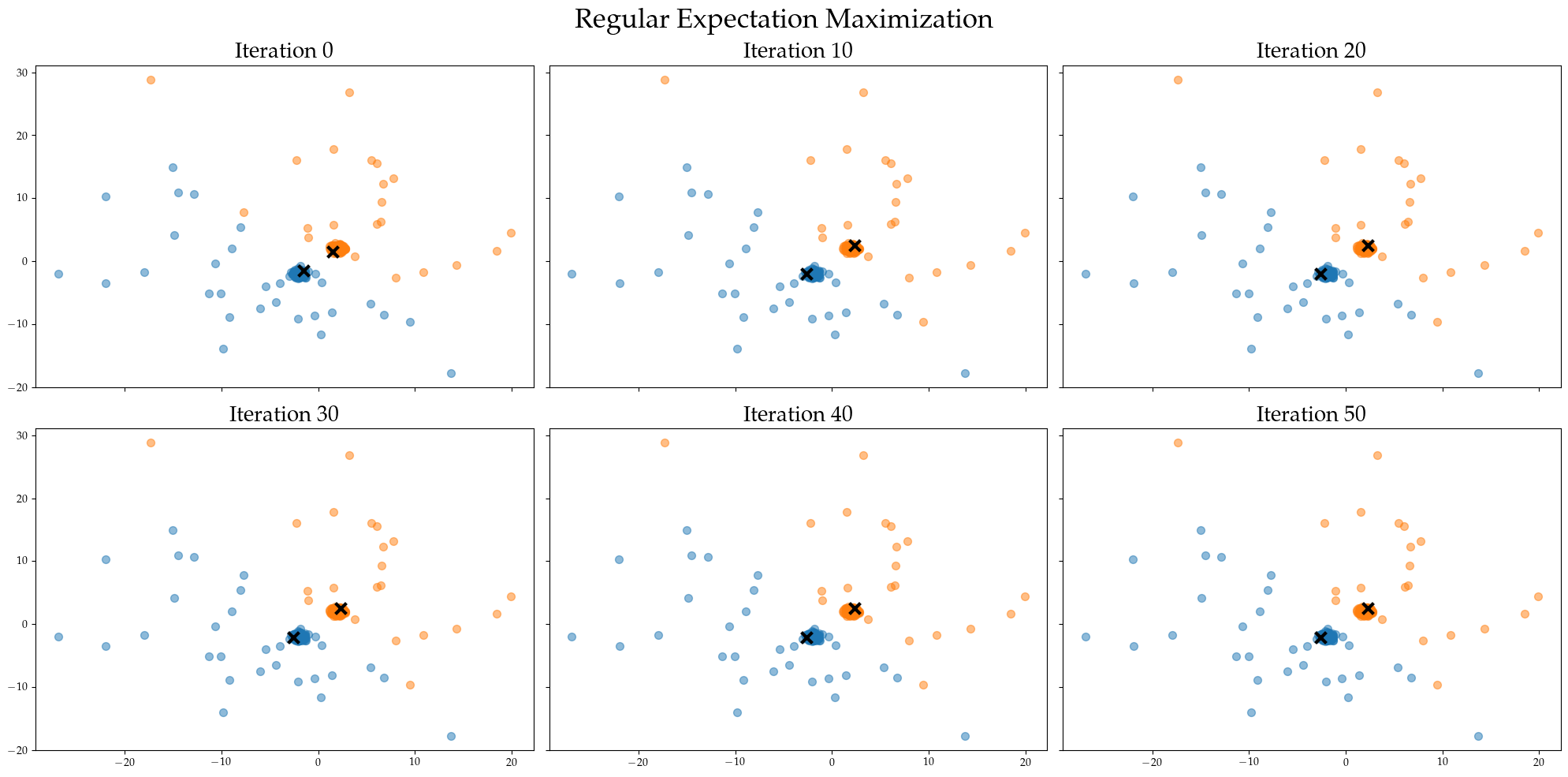}

    \caption{\textbf{Top:} With the ``smart initialization'' of the cluster centers, the points converge to the ``right'' clusters based on the original cluster centers. \textbf{Middle:} When we start the cluster centers with some shift, namely there is ``wrong'' initialization, the clusters start with good clustering but then converge to wrong solutions where they mix points with different classes. \textbf{Bottom:} For reference, we also show the regular Expectation Maximization algorithm on the shifted dataset. 
    The X's represent cluster centroids at each iteration. }
    \label{fig:expectation}
\end{figure}

To support this explanation, we also provide visualizations of the prediction space to illustrate the clustering process and the eventual drift from the training embeddings. We used a mixture of Gaussians, and trained a GMM with the Expectation Maximization algorithm using maximum likelihood, where the means are initialized based on random samples. The covariance is used as the identity matrix, with the input samples being trainable and optimizing their location. In Figure \ref{fig:expectation}, each dot represents a sample colored by its original class, where the $X$s are the centroids at each iteration. As we can see, with the ``smart initialization'' of the cluster centers, the points converge to the ``right'' clusters based on the original cluster centers. However, when we start the cluster centers with some shift, namely there is ``wrong'' initialization, the clusters start with good clustering but then converge to wrong solutions where they mix points with different classes.

\newpage

\newpage

\section{Different Parameterizations of $f$}
\label{appdx:f_abl}

In this section, we test the different ways of parameterizing the weighted-flips-to-accuracy function, $f$. Firstly, we look at the effects of not weighing each flip, and then we look at linear and cubic interpolations between flips and accuracy (as opposed to quadratic interpolation, used in the rest of the paper). Our results in Table \ref{tbl:main_abl} show that the optimal $f$ is a weighted and interpolated quadratically, with the other variations not far behind. Importantly, all variations of $f$ perform better than the second best performing method, COT \cite{lu2023predicting}.

\begin{table*}[!h]
    \centering
    \caption{Mean Absolute Error between estimated accuracy, and true accuracy on a ResNet-50 model, for weighted and unweighted flips-to-accuracy functions, that are either linear, quadratic, or cubic interpolations of points.}

    \vskip 0.15in
    \begin{tabular}{cccccc}
        \toprule
         Datasets
        & \begin{tabular}{@{}c@{}}Unweighted \\ Linear\end{tabular}
         & \begin{tabular}{@{}c@{}}Unweighted \\ Quadratic\end{tabular}
        & \begin{tabular}{@{}c@{}}Weighted \\ Linear\end{tabular}
        & \begin{tabular}{@{}c@{}}Weighted \\ Quadratic \end{tabular}
        & \begin{tabular}{@{}c@{}}Weighted \\ Cubic \end{tabular} \\
         \hline
         \multicolumn{1}{c}{\textit{Noises}} \\
         \hline
         IN-C  \{75\} \cite{hendrycks2019benchmarking} & \underline{4.95} & 5.04 & 5.94 & \textbf{4.79}  & 5.23 \\
         IN-$\overline{\mbox{C}}$ \{50\} \cite{mintun2021interaction}  & \underline{7.19} & 7.36 & 7.94 & 7.35  & \textbf{7.01} \\
         IN-3DCC \{60\} \cite{kar20223d} & \underline{4.10} & 4.12 & 4.33 & \textbf{3.66} & 4.25  \\
         CCC \{27\} \cite{press2023rdumb} & \underline{2.97} & 3.22 & 4.8 & \textbf{2.80} & 4.34 \\
         \hline
         \multicolumn{1}{c}{\textit{Domain Shifts}} \\
         \hline
        Stylized \cite{geirhos2018} & \underline{7.12} & \underline{7.12}   & \underline{7.12} & \textbf{3.81} &  \underline{7.12} \\
         IN-V2 \{3\} \cite{recht2019imagenet}  & \textbf{3.55} & \underline{3.71}  & 5.42  & 4.70 & 4.03 \\
         IN-Sketch \cite{wang2019learning} & \underline{1.11} & 1.32 & 2.64 & 4.23  & \textbf{0.23}  \\
         IN-R \cite{hendrycks2021many} & \underline{1.43} & 1.67 & 3.01 & 1.88 & \textbf{0.52} \\
          IN-D \tikzmark{start}\cite{rusak2022imagenet} &  &    & &  \\
\hspace{15mm}    Real     & 3.39  &  \underline{3.16}  &          \textbf{2.04}          & 3.18             & 4.70              \\
\hspace{15mm} Painting  & 2.07 & 1.94 &  \textbf{0.34} &    2.20       &        \underline{0.85}           \\
\hspace{15mm} Clipart  & \underline{2.78}  & 3.08       &   5.12            &                  3.37       &     \textbf{2.44}    \\
\hspace{15mm} Sketch     & \underline{6.12} & 6.95  &    12.89      &                  \textbf{5.44}       & 12.38             \\
\hspace{15mm} Infograph & \underline{7.28} & 8.76    &   10.35     &                      \textbf{3.63}     &    10.35     \\
\hspace{15mm} Quickdraw  & \textbf{0.79} & \textbf{0.79} &  \textbf{0.79} &   \underline{2.57} &  \textbf{0.79}                                    \\
         Cartoon \& Drawing \{2\} \cite{salvador2022imagenet}  & 13.60 & 13.76  & 14.34 & \underline{13.25}  & \textbf{12.96} \\

        \hline
        \multicolumn{1}{c}{\textit{Adversarial Noises}} \\
        \hline

        BG Challenge \{8\} \cite{xiao2020noise}  & \underline{7.19} & 7.36 & 7.33 &  \underline{6.92}  & 8.50 \\
        IN-A \cite{hendrycks2021nae} & 23.70 & 23.53 & \textbf{20.39} & \underline{21.61} & 22.91  \\
        IN-C Patch \{75\} \cite{gu2022evaluating}  & 1.95 & 2.00 & 2.42  & \underline{1.60} & \textbf{1.48} \\
        IN-Hard \cite{taesiri2023zoom} & 5.27 & 4.92 & \textbf{0.72}  &  3.64 & \underline{3.49} \\
        Patch-IN \{10\} \cite{pintor2023imagenet} & \textbf{7.42} &  \underline{7.55} & 9.02 &  8.87 & 7.98 \\ 
        IN-Obfuscations \{3\} \cite{stimberg2023benchmarking} & \underline{0.20} & \textbf{0.10} & \textbf{0.10} & 4.58 & \textbf{0.10} \\
        \hline
        \multicolumn{1}{c}{\textit{OOD/Other}} \\
        \hline
        ObjectNet \cite{barbu2019objectnet} & \underline{6.81} &\underline{6.81} & \underline{6.81} &  \textbf{2.74} & \underline{6.81} \\
         NINCO \cite{bitterwolf2023or} & 20.20 & 19.85 & \textbf{14.98} & 18.07 & \underline{17.73} \\
        \midrule
        \textbf{Average} & \underline{6.14} & 6.27 & 6.47 & \textbf{5.75} & 6.36 \\
        \textbf{Worst Case} & 23.70 & 23.53 & \textbf{20.39} & \underline{21.61} & 22.91 \\
        \textbf{Average (Worst Case Excluded)} & \underline{5.34} & 5.48 & 5.84 & \textbf{5.03} & 5.60 \\
    \bottomrule
    \end{tabular}

\begin{tikzpicture}[overlay, remember picture, line width=1pt]
    \draw ([yshift=+0.08cm]start.south) -- ++(0,-2.435);

    \foreach \y in {-0.15,-0.6,-1.05,-1.5,-1.9,-2.335} {
        \draw[->] ([yshift=\y cm]start.south) -- ++(1,0);
    }
\end{tikzpicture}
\label{tbl:main_abl}

\end{table*}

\newpage

\section{WF with Limited Data}
\label{appdx:wf_constraints}

To further test WF's ability in a challenging setting, we look at how it performs under memory and data constraints. To this end, we test WF in the following scenarios: (1) WF is only allowed to store 100 samples for calculating flips, and (2) when whole dataset is limited to 100 samples for flip calculation and 1,000 samples for adaptation). We note that previous work assumes the existence of at least 2,000 test samples \cite{niu2022efficient}. In both cases, we use the original weighted-flips-to-accuracy function, $f$, by multiplying the the weighted flips calculated on 100 samples by 10, and plugging the output into $f$. Even with only using 100 samples, WF is able to best the original implementation by a bit. Surprisingly, even with limited data and memory, WF manages to remain competitive with unconstrained methods, and is significantly ahead of COT, when it is constrained in a similar manner.

\begin{table*}[!h]
    \centering
    \caption{WF is effective in memory constrained settings. Without finetuning or refitting $f$, WF beats the original implementation, when only using 100 samples to calculate weighted flips (WF limited mem). In the limited memory/data setting, WF gets access to only 1000 samples in total, 100 of which are used for flip calculations. In this setting, COT gets access to 1,000 input samples and 1,000 in distribution samples. \textbf{Best} results are in bold; \underline{second best} are underlined, $\{.\}$ indicates how many splits are in each dataset, when there are more than 1.}
    
        \vskip 0.15in

        \begin{tabular}{cccccc}
        \toprule
         Datasets
         & \begin{tabular}{@{}c@{}}COT \\ original \end{tabular}
         & \begin{tabular}{@{}c@{}}WF \\ original \end{tabular}
         & \begin{tabular}{@{}c@{}}WF \\ limited mem \end{tabular}
         & \begin{tabular}{@{}c@{}}COT \\ limited  \\ mem/data \end{tabular} 
         & \begin{tabular}{@{}c@{}}WF \\ limited  \\ mem/data \end{tabular}       
         \\
         \hline
         \multicolumn{1}{c}{\textit{Noises}} \\
         \hline
         IN-C  \{75\} \cite{hendrycks2019benchmarking} & \textbf{2.23}& \underline{4.79} & 7.52 & 36.67 & 6.52\\
         IN-$\overline{\mbox{C}}$ \{50\} \cite{mintun2021interaction}  & \textbf{3.17} & 7.35 & 8.34 & 40.55 & \underline{4.60} \\
         IN-3DCC \{60\} \cite{kar20223d} & \textbf{3.02} & \underline{3.66} & 3.97 & 34.44 & 4.31 \\
         CCC \{27\} \cite{press2023rdumb} & {2.04} & {2.80} & 3.71 & 26.67 & 4.92 \\
         \hline
         \multicolumn{1}{c}{\textit{Domain Shifts}} \\
         \hline
        Stylized \cite{geirhos2018}  & {12.18} & \underline{3.81} & \underline{3.37} & 38.84 & \textbf{2.50} \\
         IN-V2 \{3\} \cite{recht2019imagenet}  & \textbf{2.68}  & 4.70 &  4.00 & 43.96 & \underline{3.80} \\
         IN-Sketch \cite{wang2019learning} & 4.23  & \underline{1.71} & \textbf{1.68} & 12.46 & 3.39\\
         IN-R \cite{hendrycks2021many} & \underline{2.44}  & \textbf{1.88} & 3.03 & 14.99 & 12.03\\
         IN-D \cite{rusak2022imagenet} &  & & & \\
\hspace{15mm}    Real       &         27.54          & \underline{3.18} & \textbf{1.73} & 41.52  & 6.51\\                          
\hspace{15mm} Painting   &         7.49          &                     \underline{2.12} & \textbf{0.71} &  26.21 & 18.44 \\                
\hspace{15mm} Clipart    &         \underline{4.52}         &                       \textbf{3.37} & 5.91 & 15.98 &  8.10 \\              
\hspace{15mm} Sketch     &          \textbf{0.71}         &                  5.44 & 6.30 & 12.65 &  \underline{4.50} \\                    
\hspace{15mm} Infograph  &           {3.44}       &                         {3.63} & \textbf{1.24}  &   4.57 & \underline{2.51} \\             
\hspace{15mm} Quickdraw  &            \underline{1.60}       &   {2.57} & 2.80 & \textbf{0.06} & 2.46 \\                                   
         Cartoon \& Drawing \{2\} \cite{salvador2022imagenet}  & \textbf{1.62} & \underline{13.25} & 16.48 & 33.25 & 13.44\\

        \hline
        \multicolumn{1}{c}{\textit{Adversarial Noises}} \\
        \hline

        BG Challenge \{8\} \cite{xiao2020noise}  & 19.68  &  \underline{6.92} & \textbf{5.84} & 32.84 & 10.15 \\
        IN-A \cite{hendrycks2021nae} & 30.38 &  {21.61} & \underline{16.75} & \textbf{15.30} & 29.15 \\
        IN-C Patch \{75\} \cite{gu2022evaluating}  & 2.57 & \textbf{1.60} & 1.98 & 47.03 & \underline{1.92} \\
        IN-Hard \cite{taesiri2023zoom} & 15.33 & \underline{3.64} & \textbf{0.65} & 5.83 & 14.73 \\
        Patch-IN \{10\} \cite{pintor2023imagenet}  &  10.13 & \textbf{8.87} & \underline{9.09} & 49.68 & 9.81 \\ 
        IN-Obfuscations \{3\} \cite{stimberg2023benchmarking}  & \underline{0.12} & {4.58} & 4.67 & \textbf{0.09} & 8.93 \\
        \hline
        \multicolumn{1}{c}{\textit{OOD/Other}} \\
        \hline
        ObjectNet \cite{barbu2019objectnet} & 10.40 & {2.74} & \textbf{0.29} & \underline{2.44} & 2.74 \\
         NINCO \cite{bitterwolf2023or} & {20.28} & \underline{18.07} & 20.24 & \textbf{13.05} & 35.68 \\
        \midrule
        \textbf{Average} & {8.17}	& \underline{5.75} & \textbf{5.67} & 23.87 & 9.40 \\
        \textbf{Worst Case} & {30.38} & \underline{21.61} & \textbf{20.24} & 49.68 &   35.68 \\
        \textbf{Average (Worst Case Excluded)} & {7.16}	& \underline{5.03} & \textbf{5.00} & 22.70 &  8.21 \\
    \bottomrule
    \end{tabular}

\begin{tikzpicture}[overlay, remember picture, line width=1pt]
    \draw ([yshift=-3.1cm]start.south) -- ++(0.0,-2.35);

    \foreach \y in {-3.275,-3.70,-4.15,-4.575,-5.025,-5.45} {
        \draw[->] ([yshift=\y cm]start.south) -- ++(1,0);
    }
\end{tikzpicture}
\label{tbl:limited}

\end{table*}

\section{Weighted Flips Ablations}
\label{appdx:iter_size_abl}

\subsection{Stopping Iteration Ablations}
\label{appdx:wf_iter}
WF measures the amount of weighted flips from iteration 0 to iteration 1,000. This is done because RDumb resets the model to its pretrained state every 1,000 iterations, in order to avoid collapse \cite{press2023rdumb}. Here, we look at how measuring weighted flips before iteration 1,000 affects the performance of WF. Interestingly, using 500 iterations increases performance by a relative 26.89\% as opposed to the 1,000 iterations used in the rest of the paper.

\begin{figure}[ht!]
\begin{minipage}{0.48\textwidth}
    \centering
    \begin{tabular}{c|c:cccc}
        \toprule
    & \multicolumn{4}{c}{Stopping Iteration} \\ 
    Datasets & 1000 & 500 & 250 & 100 & 50 \\ \hline
IN-C                     & 4.79 & 4.88 & 4.99 & 5.48 & 5.67 \\ 
IN-$\overline{\mbox{C}}$ & 7.35 & 7.48 & 7.84 & 8.96 & 10.10 \\ 
IN-3DCC                  & 3.66 & 3.32 & 3.37 & 3.00 & 3.06 \\ 
IN-V2                    & 4.70 & 4.59  & 4.93  & 5.11 & 5.49 \\ 
        \hspace{-5mm}   IN-D \begin{tikzpicture}
\draw[line width=0.5mm, ->] (-0.1,-0.2) -- (0.15,-0.2) -- (0.15,-0.45);
\end{tikzpicture} &  &   & & \\
Real                     & 3.18 & 0.36  & 0.96 & 2.82 & 5.01\\
Painting                 & 2.12 & 1.21  & 4.02 & 11.28 & 14.83\\
Clipart                  & 3.37 & 0.31  & 3.75 & 9.30 & 14.85 \\
Sketch                   & 5.44 &  2.49 & 0.33 & 5.61 & 9.26\\
Infograph                & 3.63 & 3.46  & 0.09 & 4.37 & 7.53 \\
Quickdraw                & 2.57 &  1.73 & 7.55 & 17.54 & 25.35\\
\hline
Average                  & 4.08 & \textbf{2.98} & 3.78 & 7.35 & 10.12 \\
\bottomrule
\end{tabular}

\end{minipage}
\hfill 
\begin{minipage}{0.48\textwidth}
    \centering
    \includegraphics[width=\linewidth]{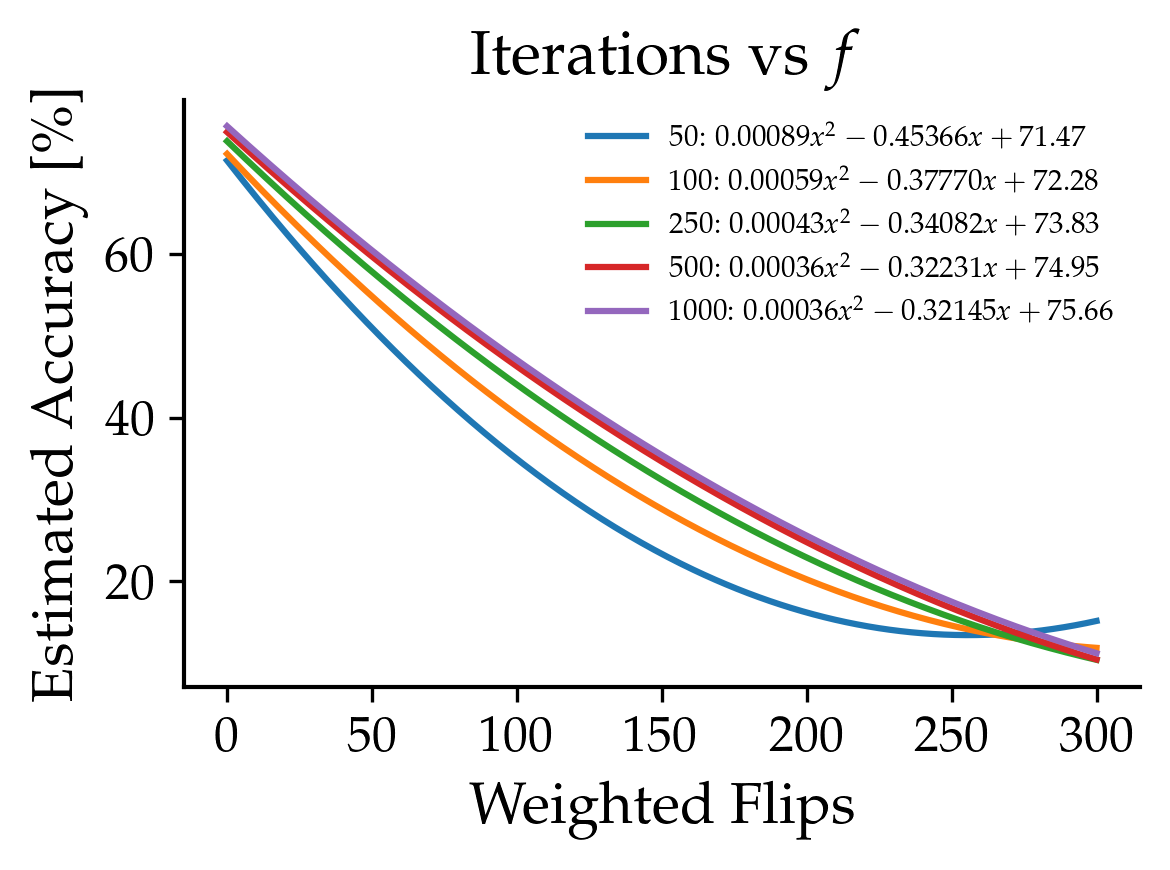}
\end{minipage}
    \vskip 0.15in
    \caption{
    \textbf{Left:} Mean Absolute Error between estimated accuracy and true accuracy, when measuring weighted flips between iteration 0 and various stopping iterations. \textbf{Right:} For different stopping iterations, interpolating between the points in the holdout set yields different weighted-flips-to-accuracy functions.}
\end{figure}
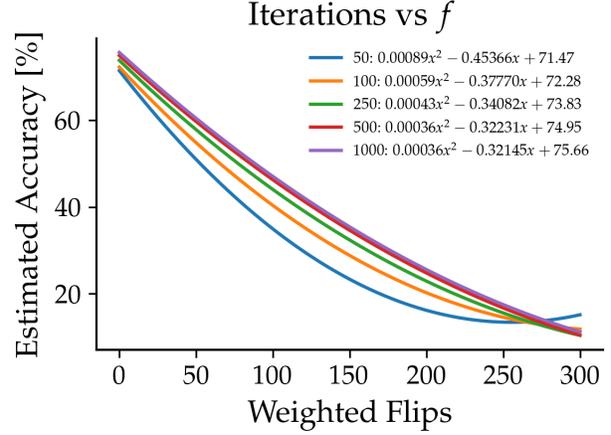

\subsection{Holdout Set Size Ablations}
\label{appdx:wf_holdout}

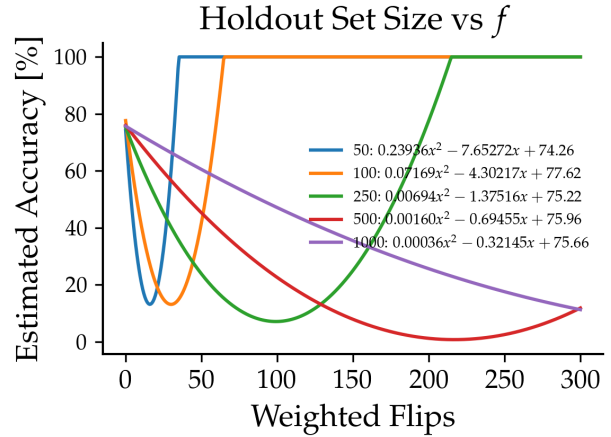
\begin{figure}[ht!]
\begin{minipage}{0.48\textwidth}

    \centering
        \begin{tabular}{c|c:cccc}
        \toprule
    & \multicolumn{4}{c}{Holdout Set Size} \\ 

    Datasets  &  1000 & 500 & 250 & 100 & 50 \\ \hline
IN-C                     & 4.79 & 4.77 & 5.52 & 6.07 & 7.62 \\ 
IN-$\overline{\mbox{C}}$  & 7.35 & 5.91 & 6.74 & 7.06 & 8.40 \\ 
IN-3DCC                   & 3.66 & 5.10 & 4.34 & 5.17 & 5.22 \\ 
IN-V2                     & 4.70 & 5.13  & 5.48  & 3.47 & 9.10 \\ 
        \hspace{-5mm}   IN-D \begin{tikzpicture}
\draw[line width=0.5mm, ->] (-0.1,-0.2) -- (0.15,-0.2) -- (0.15,-0.45);
\end{tikzpicture} &  &   & & \\
Real                      & 3.18 & 3.50  & 1.78  & 0.57 & 4.21 \\
Painting                  & 2.12 & 5.38  & 1.92  & 6.93 & 8.57 \\
Clipart                   & 3.37 & 6.69  &  7.96 & 8.79 & 7.70 \\
Sketch                    & 5.44 & 10.23 & 7.39  & 4.87  & 4.10 \\
Infograph                 & 3.63 & 6.15  & 0.88  & 3.05 & 3.77 \\
Quickdraw                 & 2.57 &  0.71 & 9.87  & 38.86   & 31.66 \\
\hline
Average                  &  \textbf{4.08} & 5.36 & 5.19 & 8.48 & 9.04 \\
\bottomrule
\end{tabular}
\label{tab:holdout_size_abl}
\end{minipage}    
\hfill 
\begin{minipage}{0.48\textwidth}
    \centering
    \includegraphics[width=\linewidth]{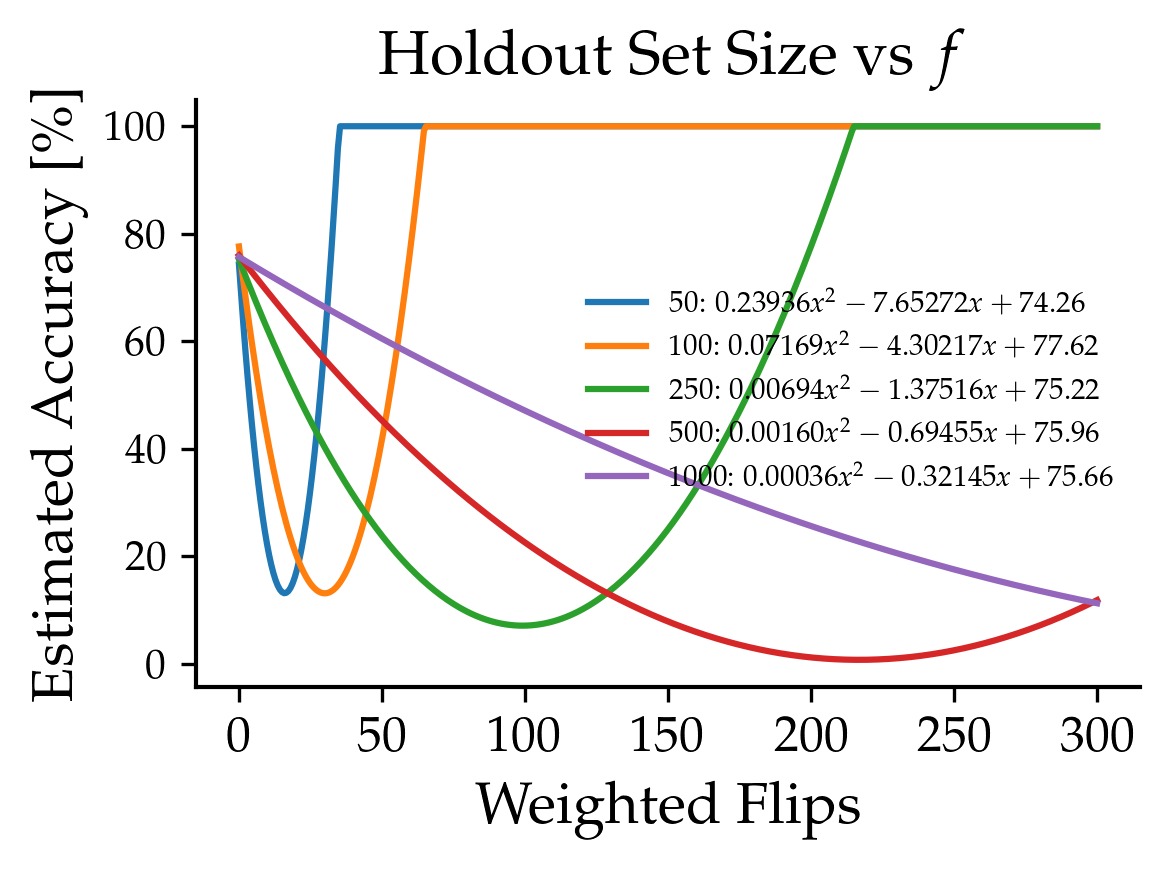}
\end{minipage}
    \vskip 0.15in
    \caption{\textbf{Left:}
    Mean Absolute Error between estimated accuracy and true accuracy, when measuring weighted flips on sets of images of different sizes. \textbf{Right:} For different holdout set sizes, interpolating between the points in the holdout set yields different weighted-flips-to-accuracy functions. $f$ can only output values that are between $0$ and $100$.}
    \vskip 0.15in
\end{figure}

\newpage

\section{WF with other TTA Methods}
\label{appdx:other_tta}

WF esimates the accuracy of a dataset as RDumb \cite{press2023rdumb} is used to adapt to it. In this section, we show that WF work with a variety of different EM methods. To further showcase the versatility of WF, we do not finetune any method, and  use the original weighted-flips-to-accuracy function $f$, for all experiments in Table \ref{tab:other_tta}. 

\begin{table*}[h] 
    \caption{
    Mean Absolute Error between estimated accuracy and true accuracy, when adapting to data using a ResNet-50 backbone and different TTA methods: Tent \cite{wang2020tent}, RPL \cite{rusak2022if}, and CPL \cite{goyal2022test}. In all cases, the original weighted-flips-to-accuracy function $f$ is used, highlighting the versatility of WF.}
    \vskip 0.15in

    \centering
        \begin{tabular}{c|c:ccc}
        \toprule
   Datasets & RDumb & Tent & RPL & CPL \\ \hline
IN-C                     & 4.79 & 6.75 & 6.85 &  5.14\\ 
IN-$\overline{\mbox{C}}$ & 7.35 & 9.68 & 7.20 & 7.44\\ 
IN-3DCC                  & 3.66 & 2.92 & 3.99 &  3.72 \\ 
IN-V2                    & 4.70 & 3.80 & 3.82 & 4.42 \\ 
        \hspace{-5mm}   IN-D \begin{tikzpicture}
\draw[line width=0.5mm, ->] (-0.1,-0.2) -- (0.15,-0.2) -- (0.15,-0.45);
\end{tikzpicture} &  &   & & \\
Real                     & 3.18 & 5.15 & 5.15 & 0.31 \\
Painting                 & 2.12 & 7.59 & 7.59 & 0.03\\
Clipart                  & 3.37 & 6.98 & 7.11 & 2.57 \\
Sketch                   & 5.44 & 3.30 & 3.52 & 3.86 \\
Infograph                & 3.63 & 2.29 & 2.24 & 3.40 \\
Quickdraw                & 2.57 & 2.24 & 2.24 & 2.37 \\
\hline
Average                  & 4.08 & 5.07 & 4.97 &  \textbf{3.33} \\
\bottomrule
        \end{tabular}
    
     \label{tab:other_tta}
\end{table*}

\section{Additional Vision Transformer Experiments}
\label{appdx:vit}

To further analyze WF and the second best method, COT, we add additionally analysis using a ViT-B/16 model. The task is to estimate the accuracy of a ViT-B/16 on a variety of datasets. We compare between using the original weighted-flips-accuracy function, $f$, which was interpolated using data from a ResNet-50, and interpolating the function using ViT-B/16 data points. In both cases, the datasets used to interpolate are the same. Additionally, we compare to COT on this task. 

\begin{table*}[h] 
    \caption{
    Mean Absolute Error between estimated accuracy and true accuracy, when estimating the accuracy of a ViT-B/16 on different datasets.}
    \vskip 0.15in

    \centering
        \begin{tabular}{c|c:ccc}
        \toprule
   Datasets & WF & WF (new $f$) & COT \\ \hline
IN-C                     & 8.34 & 1.64 & 22.24 \\ 
IN-$\overline{\mbox{C}}$ & 6.59 & 1.48 & 25.37\\ 
IN-3DCC                  & 7.19 & 1.87 & 18.43  \\ 
IN-V2                    & 4.44 & 3.63 & 21.29 \\ 
        \hspace{-5mm}   IN-D \begin{tikzpicture}
\draw[line width=0.5mm, ->] (-0.1,-0.2) -- (0.15,-0.2) -- (0.15,-0.45);
\end{tikzpicture} &  &   & & \\
Real                     & 1.02 & 7.27  & 37.21 \\
Painting                 & 3.02 & 7.06 & 19.13 \\
Clipart                  & 12.82 & 0.25 & 13.58 \\
Sketch                   & 11.04 & 1.90 & 5.74 \\
Infograph                & 9.27 & 3.34 & 1.16 \\
Quickdraw                & 2.36 & 13.04 & 0.28 \\
\hline
Average                  & 6.61 & \textbf{4.15} & 16.44 & \\
\bottomrule
        \end{tabular}
    
     \label{tab:more_vit}
\end{table*}

\newpage

\section{Omitting Samples by Top-$k$ Accuracy/Entropy Level}
\label{appdx:omitting_samples}

In addition to removing samples by Top-$k$ accuracy, we also analyze the effects of removing samples according to their initial entropy level. We find that both experiments exhibit similar behaviour: it is possible to remove many Top-$k$/low entropy samples, without significantly affecting the accuracy gain of Tent (on a holdout set of Gaussian Noise 3).

\begin{figure}[!h]
  \centering

  \begin{minipage}{0.55\linewidth}
    \centering
    \includegraphics[width=\linewidth]{figures/top_k_gain.png} %

  \end{minipage}
  \hfill %
  \begin{minipage}{0.44\linewidth}
    \centering
    \includegraphics[width=\linewidth]{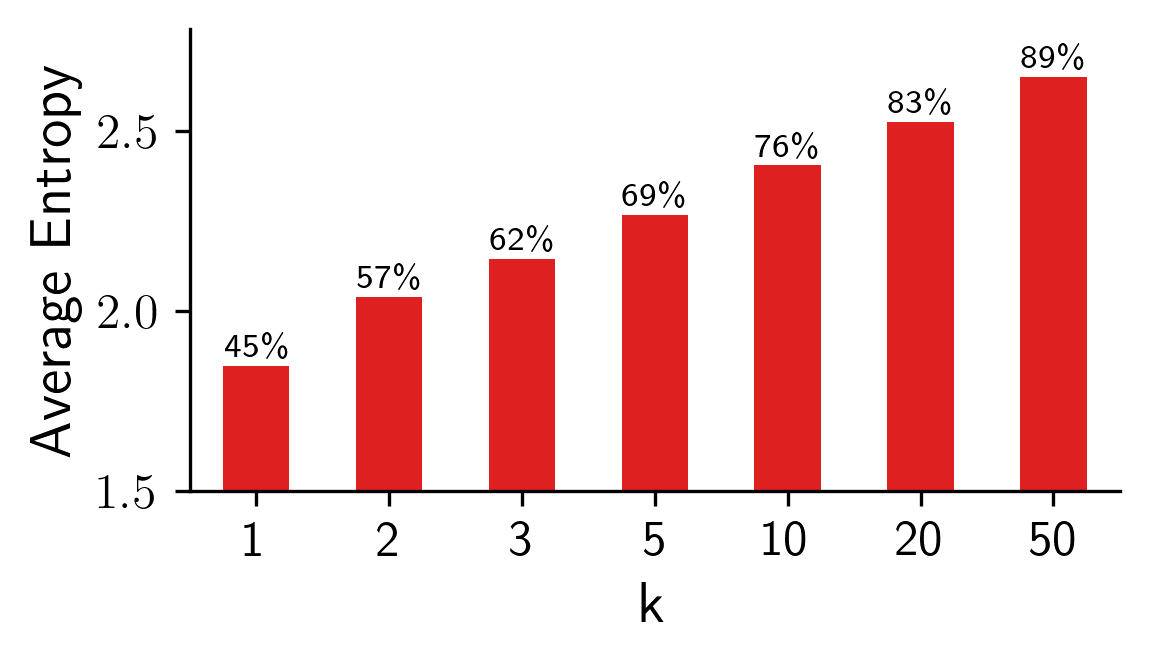} %
  \end{minipage}
      \caption{\textbf{Left:} Average entropy across top-$k$ samples for different values of $k$. The percentages shown are the fraction of images out of the whole dataset. The original dataset, Gaussian Noise 3, has an average entropy of 2.84. \textbf{Right:} The relative size of the datasets, when top-$k$ samples are removed.}
\end{figure}

\begin{figure}[!h]
  \centering
  \begin{minipage}{0.55\linewidth}
    \centering
    \includegraphics[width=\linewidth]{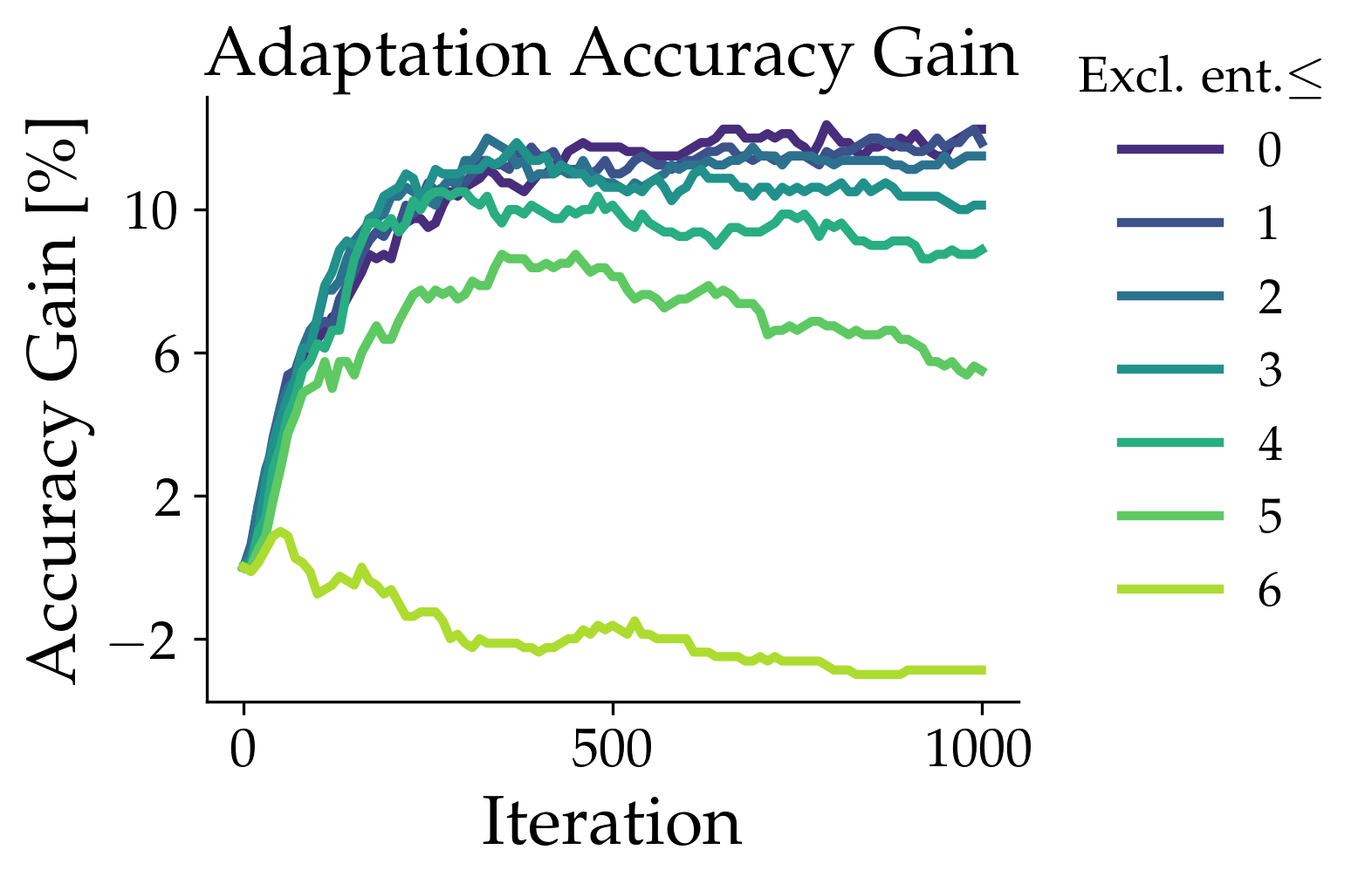} %

  \end{minipage}
  \hfill %
  \begin{minipage}{0.44\linewidth}
    \centering
    \includegraphics[width=\linewidth]{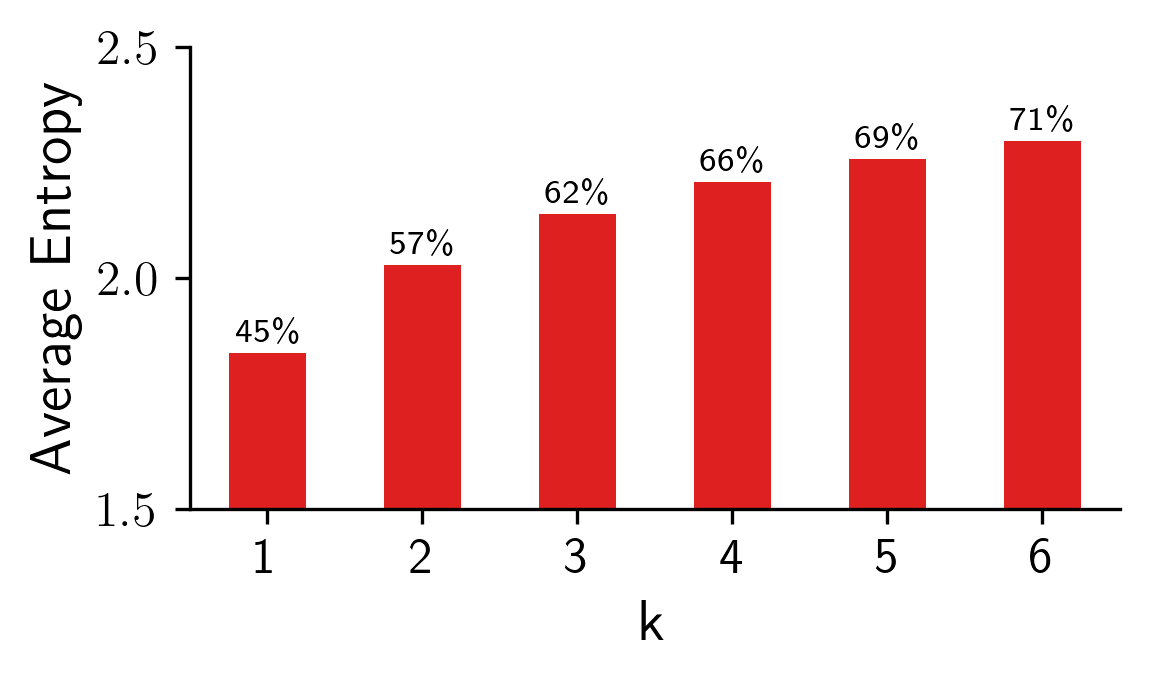} %
  \end{minipage}
      \caption{\textbf{Left:} Accuracy gain per iteration on a holdout set, as Tent adapts to its inputs. Each line corresponds to a different experiment where we remove samples based on their initial entropy level. Similarly to Figure \ref{fig:topk}, it's possible to remove low entropy samples while barely hurting performance. When entropy $\leq0$, no images are excluded. \textbf{Right:} The relative size of the datasets and their average entropy, when samples with a entropy level $\leq k$ are removed.}
\end{figure}

\newpage

\section{Silhouette score, Shift distance, and Accuracy Throughout Entropy Minimization}
\label{appdx:sil_shift}

In Figure \ref{fig:silhouette}, we looked at the changes of Silhouette scores/Shift distances for each phase in EM. Here, we show how these scores, along with accuracy, change in every iteration of Tent. For each one of the datasets analyzed, we group noises based on severity level, and plot their averages and standard deviations, for every iteration.

\begin{figure}[!h]
  \centering
  \includegraphics[width=0.99\linewidth]{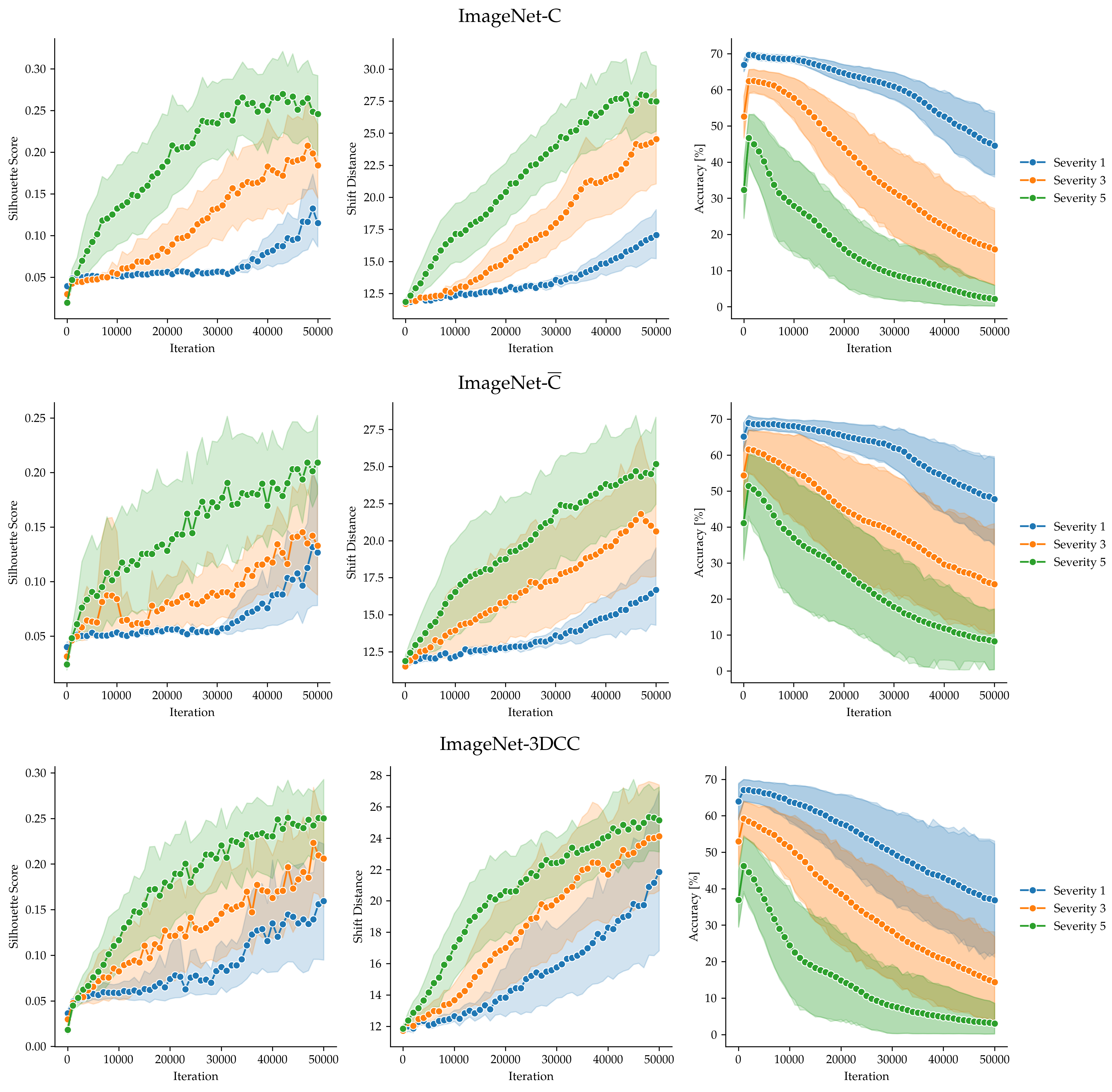}
  \caption{Changes in Silhouette scores, Shift distances, and Accuracies as Tent adapts to its inputs. We group together noises by severity level, and average the data for every iteration.}

  \label{fig:sil_shift_acc}
\end{figure}

\newpage

\section{WF on CIFAR10/100}
\label{appdx:cifar}

WF is not as effecitve on CIFAR10 \cite{krizhevsky2009learning} as it is on ImageNet \cite{deng2009imagenet} scale datasets. CIFAR10 is an outlier in entropy minimization: for example, \citet{press2023rdumb} showed that Tent doesn’t degrade in accuracy, even after 100 million CIFAR10 images seen. We nonetheless run our method on CIFAR10. On average, we see only 0-5 label flips per dataset on C10-C. This is far from what we see ImageNet-scale datasets we tested.

Like in the paper, we interpolate a weighted-flips-to-accuracy function $f$ on the holdout set and get:

$$f(x) = -249.36x^2 - 87.39x + 77.01$$

which has a MAE of 16.64 on the C10 validation set. 

We repeat this for CIFAR100 and get:

$$f(x) = 0.000322x^2 - 0.287x + 99.54$$

which has a MAE of 9.10 on the C100 validation set.

Apart from refitting $f$, we did not tune any other parameter in these two experiments.

\section{RDumb}
\label{appdx:rdumb}

WF uses RDumb \cite{press2023rdumb} to estimate accuracy. We go over the implementation of the method in brief. RDumb is based on ETA \cite{niu2022efficient}, wherein the model is reset to its pretrained state every 1,000 iterations. Rdumb optimizes the BatchNorm \cite{ioffe2015batch} parameters, $\Theta$ of a given classifier $f$.

The loss optimized is entropy, with two filtration steps: the first, in which samples with high entropy are filtered out, and the second, in which samples that produce logits similar to previous samples are filtered out. 

For a sample $x$, the first filtration is given by:

$$S^{ent}(x) = \frac{1}{\exp[E(x; \Theta) - E_0]} \cdot \mathbb{I}_{E(x;\Theta) < E_0}(x),$$

with $E_0= 0.4 \times \text{ln} 10^3$.

The second filtration is given by:

$$S^{div}(x) = \mathbb{I}_{\{cos(f_o(x), m^{-1}) < \epsilon\}}(x)$$

where $cos()$ is the cosine similarity, and $m^t$ is an exponential moving average of the logits of previously seen samples at iteration $t$:

$$m^{t} = 
\begin{cases} 
y^{1}, & \text{if } t = 1 \\
\alpha y^{t} + (1 - \alpha)m^{t-1}, & \text{if } t > 1 
\end{cases}
$$

and $y^{t}$ is the average model prediction on a batch of inputs at step $t$, and $\alpha = 0.9$.

Put together with entropy minimization, the optimization formula becomes:

$$\min_{\hat{\Theta}} - S^{ent}(x) \cdot S^{div}(x) \sum_{y \in C} f_{\Theta}(y|x)\log f_{\Theta}(y|x)$$

RDumb uses a SGD with a learning rate of $2.5\times10^{-4}$, and a batch size of 64, and is reset to its pre-trained state every 1,000 iterations.

\section{Software Licenses}

\begin{itemize}
    \item ImageNet-C \citep{hendrycks2019benchmarking}Apache License 2.0\\
        \url{https://github.com/hendrycks/robustness}
    \item ImageNet-R \citep{hendrycks2021many} MIT License\\
        \url{https://github.com/hendrycks/imagenet-r}
    \item ImageNet-3D-CC \citep{kar20223d}: CC-BY-NC 4.0 License\\
    \url{https://github.com/EPFL-VILAB/3DCommonCorruptions}
    
    \item ImageNet-$\overline{\mbox{C}}$ \citep{mintun2021interaction}: MIT License\\
    \url{https://github.com/facebookresearch/augmentation-corruption}

    \item ImageNet-V2 \citep{recht2019imagenet}: MIT License\\
    \url{https://github.com/modestyachts/ImageNetV2}

    \item Backgrounds Challenge \citep{xiao2020noise}: \\
    \url{https://github.com/MadryLab/backgrounds_challenge}
    
    \item CCC \citep{press2023rdumb}: MIT License\\
    \url{https://github.com/oripress/CCC}
    \item Stylized ImageNet
    \cite{geirhos2018}: MIT License \\
    \url{https://github.com/rgeirhos/Stylized-ImageNet}
    \item NINCO
    \cite{bitterwolf2023or}: MIT License
    \url{https://github.com/j-cb/NINCO}
    \item ImageNet-D
    \cite{rusak2022imagenet}: Apache License 2.0 \\
    \url{https://github.com/bethgelab/robustness}
    \item ObjectNet 
    \cite{barbu2019objectnet}: MIT License
    \url{https://objectnet.dev/}
    \item Shift Happens Benchmark: Apache License 2.0
    \url{https://github.com/shift-happens-benchmark/icml-2022}
\end{itemize}

\end{document}